\PassOptionsToPackage{unicode}{hyperref}
\PassOptionsToPackage{hyphens}{url}
\documentclass[
  11pt,
]{article}
\usepackage{xcolor}
\usepackage[margin=1in]{geometry}
\usepackage{amsmath,amssymb}
\usepackage{iftex}
\ifPDFTeX
  \usepackage[T1]{fontenc}
  \usepackage[utf8]{inputenc}
  \usepackage{textcomp} % provide euro and other symbols
\else % if luatex or xetex
  \usepackage{unicode-math} % this also loads fontspec
  \defaultfontfeatures{Scale=MatchLowercase}
  \defaultfontfeatures[\rmfamily]{Ligatures=TeX,Scale=1}
\fi
\usepackage{lmodern}
\ifPDFTeX\else
\fi
\IfFileExists{upquote.sty}{\usepackage{upquote}}{}
\IfFileExists{microtype.sty}{% use microtype if available
  \usepackage[]{microtype}
  \UseMicrotypeSet[protrusion]{basicmath} % disable protrusion for tt fonts
}{}
\makeatletter
\@ifundefined{KOMAClassName}{% if non-KOMA class
  \IfFileExists{parskip.sty}{%
    \usepackage{parskip}
  }{% else
    \setlength{\parindent}{0pt}
    \setlength{\parskip}{6pt plus 2pt minus 1pt}}
}{% if KOMA class
  \KOMAoptions{parskip=half}}
\makeatother
\usepackage{longtable,booktabs,array}
\usepackage{calc} % for calculating minipage widths
\usepackage{etoolbox}
\makeatletter
\patchcmd\longtable{\par}{\if@noskipsec\mbox{}\fi\par}{}{}
\makeatother
\IfFileExists{footnotehyper.sty}{\usepackage{footnotehyper}}{\usepackage{footnote}}
\makesavenoteenv{longtable}
\usepackage{graphicx}
\makeatletter
\newsavebox\pandoc@box
\newcommand*\pandocbounded[1]{% scales image to fit in text height/width
  \sbox\pandoc@box{#1}%
  \Gscale@div\@tempa{\textheight}{\dimexpr\ht\pandoc@box+\dp\pandoc@box\relax}%
  \Gscale@div\@tempb{\linewidth}{\wd\pandoc@box}%
  \ifdim\@tempb\p@<\@tempa\p@\let\@tempa\@tempb\fi% select the smaller of both
  \ifdim\@tempa\p@<\p@\scalebox{\@tempa}{\usebox\pandoc@box}%
  \else\usebox{\pandoc@box}%
  \fi%
}
\def\fps@figure{htbp}
\makeatother
\providecommand{\tightlist}{%
  \setlength{\itemsep}{0pt}\setlength{\parskip}{0pt}}
\usepackage{bookmark}
\IfFileExists{xurl.sty}{\usepackage{xurl}}{} % add URL line breaks if available
\hypersetup{
  pdftitle={When Agent Governance Helps},
  pdfauthor={Michael Ray Johnson (MJ); Linda Naimi},
  hidelinks,
  pdfcreator={LaTeX via pandoc}}

\title{When Agent Governance Helps}
\author{Michael Ray Johnson (MJ)\\[2pt] Department of Technology, Leadership, and Innovation,\\ Purdue University, West Lafayette, Indiana, USA\\ \texttt{john3678@purdue.edu} \quad \texttt{johnson.michael@gmail.com}\\[1pt] ORCID 0009-0008-8884-0587 \and Linda Naimi\\[2pt] Department of Technology, Leadership, and Innovation,\\ Purdue University, West Lafayette, Indiana, USA\\ \texttt{lnaimi@purdue.edu}}
\date{}

\usepackage{xurl}
\begin{document}
\maketitle

\begin{abstract}
No specification says how a governed autotelic AI agent organization,
where agents pursue self-generated goals inside guardrails, should be
designed and evaluated. We answer in two parts. First, we synthesize the
Governed Autotelic Multi-Agent Product Organization (GAMPO) framework
from a document-based qualitative evidence synthesis of 321 sources,
integrating agency, agile, platform, and governance theory into a
runnable specification. Second, we probe a prompt-layer instantiation of
GAMPO on CHI-Bench, a long-horizon healthcare benchmark, across open and
frontier models. The result is a boundary condition: governance benefit
is gated by a model's spare capacity and is domain- and model-specific.
On capacity-constrained open models the full procedure yields no
reliable benefit, whereas a single ``verify your writes'' sentence
doubles task success (pass@1 2/20 to 4/20). At the frontier the same
scaffold lifts prior-authorization 24\% to 40\% but nets zero on another
model, a gap traced to a stable recommendation-override disposition. A
second result refines the first: replacing the generic procedure with an
answer-blind, per-task definition-of-done, keyed only to the case's own
policy and published standards and never the hidden key, raises
prior-authorization to 84\% under best-of-five self-consistency
selection (the stricter single-attempt figure is lower, now confirmed at
68\% by a held-out board) and utilization-management to 44\%, while
care-management meets a subjective content-quality wall. The
contribution is a named, auditable framework and capability-gated
evidence that governance should be sized to spare capacity, and that at
the frontier a case-grounded specification beats a uniform procedure.
Findings are exploratory: partial instantiation, small per-cell samples
(n = 5-25), single trials, and advisory numbers separating best-of-five
self-consistency from single-attempt pass@1.
\end{abstract}


\begin{figure}
\centering
\includegraphics[width=1\linewidth,height=0.78\textheight,keepaspectratio,alt={Figure 1. Graphical abstract of the paper's two central findings. Panel 1 (capability gating): prepending a comprehensive governance procedure helps only in proportion to a model's spare reasoning capacity, below the capability floor the effect is null, on capacity-constrained models the full procedure is neutral or harmful (one 35B model loses 0.11 on prior-authorization) while a single ``verify your writes'' sentence doubles pass@1 (2 to 4 of 20), and at the frontier it helps where the model has traction (Opus 4.8). Panel 2 (the form of governance): the bars give the full progression from the unassisted baseline (no GAMPO) through the uniform procedure to the answer-blind, per-task definition-of-done, prior-authorization 24\% to 40\% to 84\% and utilization-management 8\% to 16\% to 44\% (the black markers give the stricter single-attempt pass@1 beneath the best-of-five self-consistency advisory bars (utilization-management confirmed at 39.2\%, prior-authorization now confirmed at 68\% by a held-out single-run board)), while care-management stalls at a subjective-judge content-quality wall. The advisory is answer-blind: a specification keyed only to the case's own policy and to published United States standards (the CMS WISeR prior-authorization model; the CMS-0057-F Provider-Access and Payer-to-Payer application programming interfaces; the CCM, APCM, and GUIDE care-management programs; CMS and AMA coding with chart-documentation fidelity; and NCD/LCD and NASS/InterQual clinical criteria), never the hidden answer key. Panel 1 is a conceptual schematic; precise effects appear in Figures 8, 9, and 11 and Tables 1--6. Source: Sections 5.2--5.6.}]{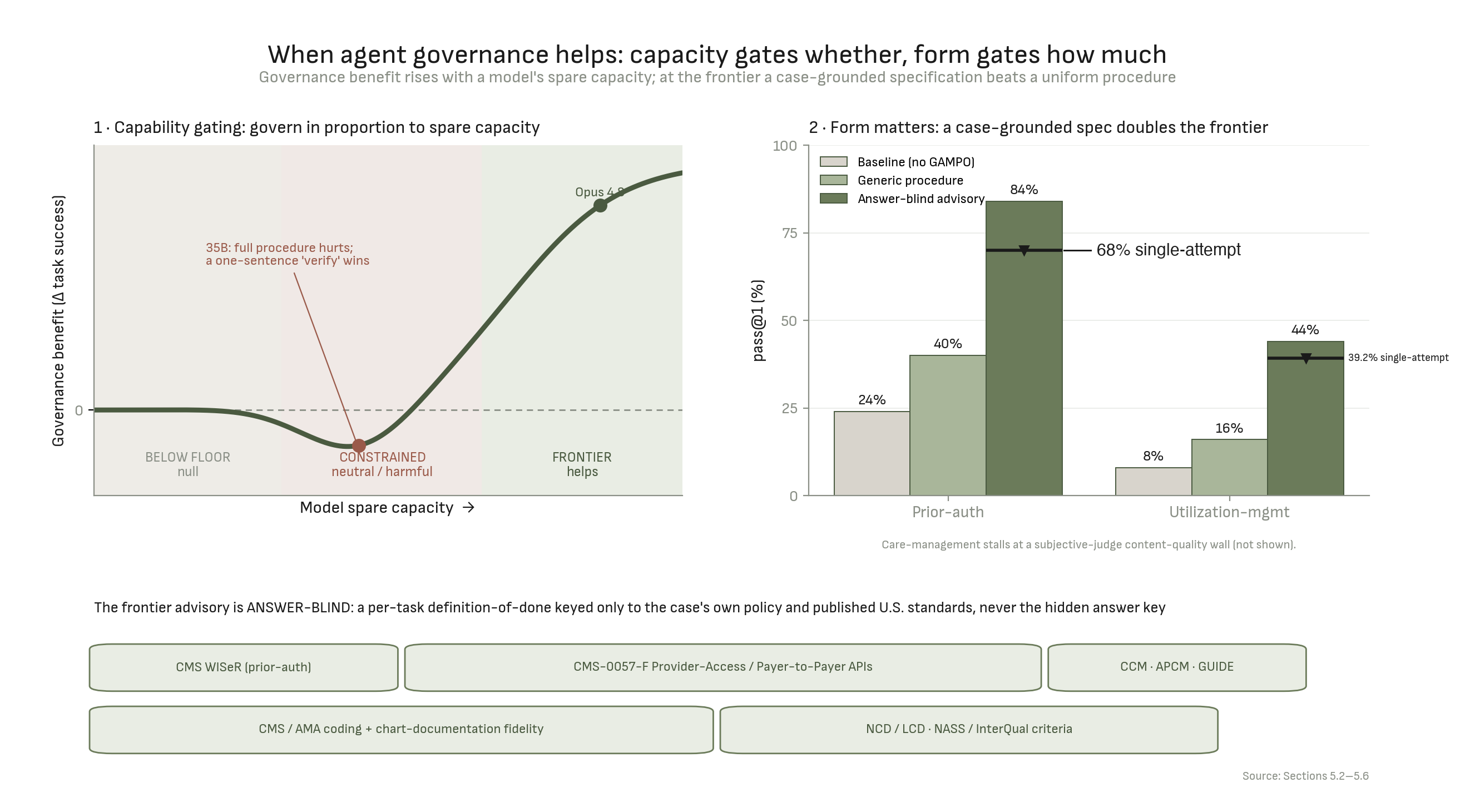}
\caption{Graphical abstract of the paper's two central
findings. Panel 1 (capability gating): prepending a comprehensive
governance procedure helps only in proportion to a model's spare
reasoning capacity, below the capability floor the effect is null, on
capacity-constrained models the full procedure is neutral or harmful
(one 35B model loses 0.11 on prior-authorization) while a single
``verify your writes'' sentence doubles pass@1 (2 to 4 of 20), and at
the frontier it helps where the model has traction (Opus 4.8). Panel 2
(the form of governance): the bars give the full progression from the
unassisted baseline (no GAMPO) through the uniform procedure to the
answer-blind, per-task definition-of-done, prior-authorization 24\% to
40\% to 84\% and utilization-management 8\% to 16\% to 44\% (the black
markers give the stricter single-attempt pass@1 beneath the best-of-five
self-consistency advisory bars (utilization-management confirmed at
39.2\%, prior-authorization now confirmed at 68\% by a held-out
single-run board)), while care-management stalls at a subjective-judge
content-quality wall. The advisory is answer-blind: a specification
keyed only to the case's own policy and to published United States
standards (the CMS WISeR prior-authorization model; the CMS-0057-F
Provider-Access and Payer-to-Payer application programming interfaces;
the CCM, APCM, and GUIDE care-management programs; CMS and AMA coding
with chart-documentation fidelity; and NCD/LCD and NASS/InterQual
clinical criteria), never the hidden answer key. Panel 1 is a conceptual
schematic; precise effects appear in Figures 8, 9, and 11 and Tables
1--6. Source: Sections 5.2--5.6.}
\end{figure}

\section{Summary of Contributions}\label{summary-of-contributions}

\textbf{Purpose.} Organizations are beginning to delegate product and
software work to teams of AI agents that set their own subgoals. Leaders
need to know two things: what a defensible governance design looks like,
and whether adding governance to an agent actually improves outcomes.
This study answers both, and the answers are not the same.

\textbf{What changed because of this study.} We deliver a named
framework (GAMPO) synthesized from 321 appraised sources, and we
contribute capability-gated evidence on agent governance: prepending a
comprehensive governance procedure to an agent helps only when the model
has spare reasoning capacity to spend on it. On weaker models the same
procedure is at best neutral and occasionally harmful, while a
one-sentence verification habit helped consistently across our runs. At
the frontier, the form of governance matters as much as its presence:
replacing the uniform procedure with an answer-blind, per-task
definition-of-done raised prior-authorization from 40\% to 68\%
single-attempt (84\% under best-of-five self-consistency).

\textbf{Decision implications.} Govern in proportion to spare capacity.
For ordinary, capacity-constrained deployments, invest first in cheap,
action-grounded checks (verify that writes persisted) rather than in
long procedural scaffolds; reserve comprehensive governance for frontier
models, and even there prefer a case-grounded, answer-blind
specification of done over a longer uniform procedure, expecting domain-
and model-specific effects rather than a uniform lift. These are
associational findings from small samples; treat them as design
heuristics, not causal guarantees.

\textbf{Evidence snapshot.} (1) Framework evidence base: n = 321 sources
(74.8\% peer-reviewed), appraised with AMSTAR 2, CASP, JBI, and AACODS.
(2) Benchmark: CHI-Bench, 75 tasks across three healthcare domains,
fractional reward in {[}0,1{]} plus binary pass@1. (3)
Capacity-constrained model: the pooled fractional lift 0.431 to 0.590
and the pass@1 doubling 2/20 to 4/20 came from a one-sentence verify
prompt, not the full framework, which was net-negative (pooled 0.393).
(4) Frontier prior-authorization lift replicates across vendors: paired
pass@1 +16 points on Opus 4.8 (24\% to 40\%) and +20 points (12\% to
32\%, fractional +0.056) on each of two 2026-generation models (Google's
Gemini 3.1 Pro, Moonshot's Kimi 2.7-code); the same procedure is
neutral-to-harmful outside prior-authorization (Gemini
utilization-management fractional Delta -0.248, Kimi care-management
-0.188) and near zero on some frontier models (Fable 5, GPT-5.6-sol).
(5) Filed leaderboard submission (full GAMPO on Opus 4.8): 37.3\%
overall (28/75; Wilson 95\% CI 27--49), led by care-management at 56\%
(95\% CI 37--73), prior-authorization 40\% (95\% CI 23--59),
utilization-management 16\% (95\% CI 6--35). (6) Answer-blind per-task
advisory (a case-grounded definition-of-done, never reading the hidden
answer key) roughly doubles the frontier prior-authorization result over
the generic procedure, 40\% to 84\% pass@1 under best-of-five
self-consistency (strict single-attempt 68\%, 17 of 25, now confirmed by
a held-out single-run board), via fidelity to chart-documented codes;
utilization-management reaches 44\% self-consistency / 39.2\%
single-attempt; care-management stalls at a subjective judge
content-quality wall (the grounded content advisory reaches 40\%
single-draw, 10 of 25; a sharpened 52\% round was overturned as
redistribution under simulated-member consent variance), and
care-management is the one domain where the generic procedure (56\%)
beats the content advisory. (7) Single-trial (and, for the advisory
boards, single-draw or best-of-five) design; per-cell n = 5--25; results
exploratory (pass@1 confidence intervals are wide and overlap, so read
rankings as indicative; self-consistency figures are not the
leaderboard's single-attempt metric). (8) Implementation status: the
framework is instantiated in a runnable sibling repository
(gampo-runtime); the reasoning and governance core (nine capability
servers running the ten-capability cycle, operating cycle, constraint
and promotion gates, audit ledger, recursive self-improvement loop) is
fully implemented, integration surfaces are answer-blind simulations,
and the empirical findings above exercise only the prompt-layer kernel
(Figure 12 versus Figure 5). (9) Cross-generation replication (not a
filed submission): the same governance stack re-run on Opus 5,
single-agent and strict single-attempt, with the answer-blind advisory
on prior-authorization and utilization-management and the generic
procedure only on care-management, scores prior-authorization 72\%
(18/25), utilization-management 36\% (9/25), and care-management 56\%
(14/25), 54.7\% overall (41/75), reproducing the Opus 4.8 domain pattern
on a newer frontier generation (single trial per task, same 25 public
tasks per domain).

\section{1. Introduction}\label{introduction}

\textbf{Context.} A \emph{multi-agent system} is software in which
several autonomous agents interact to accomplish work (Wooldridge,
2009). When the agents are large language models orchestrated to plan,
write, test, and ship code, the system begins to resemble a small
product organization that runs with little human supervision. The most
ambitious version of this idea is \emph{autotelic}: an agent that
``generates, scores, and pursues its own bounded goals within governance
constraints,'' extending the psychological notion of intrinsically
motivated, self-contained activity (Csikszentmihalyi, 1990) and its
computational formalization as intrinsically motivated goal-conditioned
learning (Colas et al., 2022). The promise is an organization that
starts lean and improves itself; the peril is ungoverned autonomy in
consequential settings. Standards and scholarship now specify
\emph{what} such systems must satisfy, from risk-management functions
(NIST, 2023; ISO/IEC, 2023) to foundation-model social risks (Bommasani
et al., 2021), but say comparatively little about \emph{how} to operate
an agent organization day to day.

\textbf{Problem statement.} No integrative framework specifies how a
governed autotelic multi-agent product organization should be designed,
operated, and evaluated as an applied technology system, and,
separately, the field lacks evidence on a prior question: does adding
governance scaffolding to a capable agent actually improve its behavior,
or only constrain it? The two gaps are connected. A governance
specification that has never been instrumented against a hard task is an
assertion; a benchmark result with no specification behind it is a
trick. This paper supplies both halves and shows that their interaction
is itself the finding.

\textbf{Contribution.}

\begin{itemize}
\tightlist
\item
  \emph{Theoretical.} A named, integrative framework, the Governed
  Autotelic Multi-Agent Product Organization (GAMPO), that combines
  autotelic-agency, tacit-knowledge, disciplined-agile, platform, and
  sociotechnical-governance theory into one coherent design vocabulary
  under a pragmatist anchor.
\item
  \emph{Empirical.} Capability-gated evidence on agent-governance
  scaffolding: governance benefit is gated by a model's spare capacity
  and is domain- and model-specific, with a one-sentence verification
  prompt dominating the full framework on capacity-constrained models.
\item
  \emph{Practical/executive.} A decision rule, govern in proportion to
  spare capacity, and an autonomy ladder with non-delegable boundaries
  that translate the framework into an operating posture.
\item
  \emph{Methodological.} A demonstration that qualitative evidence
  synthesis (QES) plus a single prompt-layer benchmark probe can convert
  fragmented secondary literature into an auditable specification and
  then stress-test its central claim.
\end{itemize}

\textbf{Paper roadmap.} Section 2 synthesizes the literature and names
the framework. Section 3 states the research questions and propositions.
Section 4 describes the two-phase synthesis method and the benchmark
probe. Section 5 reports results, the framework's structure and the
benchmark's capability-gated findings. Section 6 interprets them for
executives and academics. Section 7 states limitations and a
future-research agenda. Section 8 concludes.

\section{2. Background and Literature
Synthesis}\label{background-and-literature-synthesis}

\textbf{Search strategy and source base.} The framework was built by a
document-based qualitative evidence synthesis that updated and
integrated the agentic-AI and governance literatures. Searches spanned
eight academic databases plus arXiv/SSRN, practitioner publications,
standards bodies, and vendor documentation, over January 2018 to Q1
2026. A PRISMA 2020-aligned process (Page et al., 2021) moved from 1,847
records identified to 1,234 screened, 612 sought for retrieval, 421
assessed in full text, and \textbf{321 included} (the full selection
flow appears in Figure 3, Section 4). Sources were appraised by type
with four instruments, AMSTAR 2 for systematic reviews (Shea et al.,
2017), CASP for qualitative studies (Critical Appraisal Skills
Programme, 2018), JBI for quantitative studies (Aromataris \& Munn,
2020), and AACODS for grey literature (Tyndall, 2010), and classified
Include / Weight / Exclude in a public evidence ledger.

\textbf{Known evidence.} Three findings anchor the design. First,
agentic coding tools have advanced quickly but with weak integrated
governance, and their gains are workflow-dependent rather than
universal: a controlled study found experienced developers on mature
codebases were about 19\% \emph{slower} with AI assistance (Becker et
al., 2025), reconciling with mixed field data and cautioning against
context-free productivity claims. Second, successful agent deployments
are overwhelmingly hybrid: roughly 70\% of effective implementations
keep humans in the loop from the start (Bornet et al., 2025). Third,
governance scholarship is rich on management-system requirements (NIST,
2023; ISO/IEC, 2023) and emerging operational patterns such as
Governance-as-a-Service with graded policy modes (Gaurav et al., 2025),
but thin on a runnable, repository-level operating model. The consensus
is that governance matters; the unresolved debate is \emph{how much, of
what kind, and when it pays off}.

\textbf{Gap statement.} Precisely: there is no auditable, tool-agnostic
specification that tells an agent organization which artifacts to keep,
which actions never to delegate, and how to evaluate itself, and there
is no empirical test of whether instantiating such a specification
improves a capable agent's task behavior. The gap is a population (agent
product organizations), a measure (does governance scaffolding change
task success), and a mechanism (under what model conditions).

\textbf{Conceptual frame.} GAMPO integrates six traditions: autotelic
agency for the goal engine (Colas et al., 2022; Csikszentmihalyi, 1990);
tacit/explicit knowledge for an epistemic taxonomy that separates
\emph{deterministic} claims (truth fixed by rule or test) from
\emph{probabilistic} claims (Polanyi, 1966; Nonaka \& Takeuchi, 1995);
Disciplined Agile for context-driven way-of-working selection (Ambler \&
Lines, 2022); platform theory for spin-off and option logic (Parker et
al., 2016; Tiwana, 2014); sociotechnical AI governance for non-delegable
boundaries and an audit ledger (NIST, 2023; ISO/IEC, 2023); and Schank's
(2023) Process Inventory Framework as an organizing coverage map. The
framework predicts that GAMPO outcomes, a vector of usefulness, safety,
trust, simplicity, and business value, rise with governance scaffolding,
epistemic classification, topology leanness, experience observability,
and non-delegable-boundary discipline, moderated by way-of-working fit
and economic routing (Figure 2).

\begin{figure}
\centering
\includegraphics[width=1\linewidth,height=0.78\textheight,keepaspectratio,alt={Figure 2. Theoretical framework: latent design constructs (governance scaffolding, epistemic classification, topology leanness, experience observability, boundary discipline) predicting the GAMPO outcome vector, with moderators. Source: synthesis of n = 321 appraised sources; conceptual, not estimated.}]{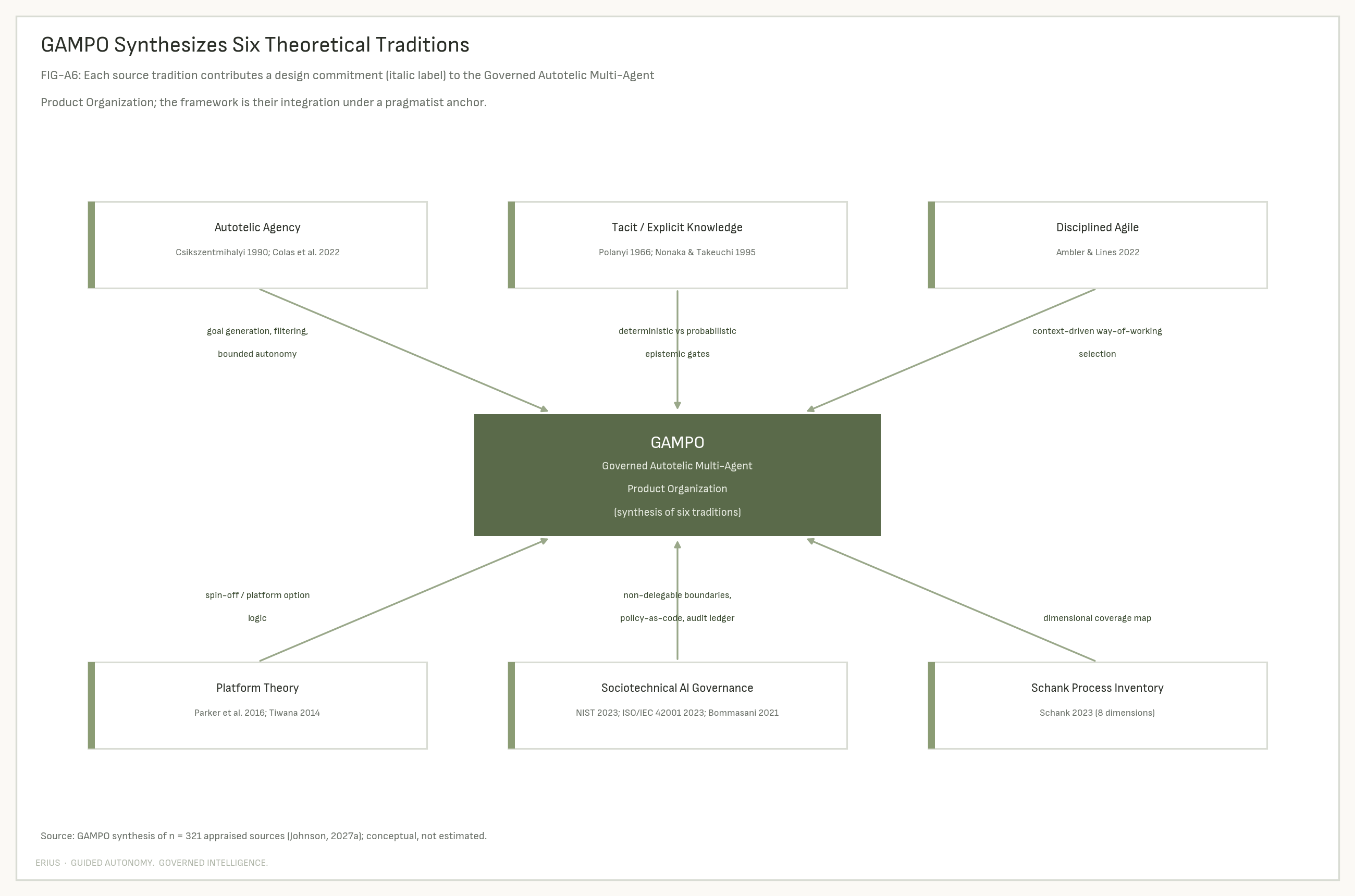}
\caption{Theoretical framework: latent design constructs
(governance scaffolding, epistemic classification, topology leanness,
experience observability, boundary discipline) predicting the GAMPO
outcome vector, with moderators. Source: synthesis of n = 321 appraised
sources; conceptual, not estimated.}
\end{figure}

\section{3. Research Questions and
Propositions}\label{research-questions-and-propositions}

\textbf{Primary research question.} How can a governed autotelic
multi-agent AI product organization be designed, governed, and evaluated
as an applied technology system that starts lean, generates and
evaluates its own bounded goals, and produces auditable evidence of
usefulness, safety, trust, and business value?

\textbf{Secondary research questions.}

\begin{itemize}
\tightlist
\item
  \emph{RQ1 (governance and architecture).} What relationships between
  governance scaffolding, architectural boundaries, and observable agent
  behavior are supported by the synthesized evidence, and how should
  those scaffolds be designed?
\item
  \emph{RQ2 (economic and process efficiency).} What evidence supports
  differences between lean and bloated multi-agent topologies, ways of
  working, and model-routing strategies on productivity, quality, cost,
  and learning velocity?
\item
  \emph{RQ3 (evaluative methodology).} What benchmarking and evaluation
  methodology generates auditable, longitudinal, tool-agnostic evidence
  of usefulness, safety, trust, and business value?
\end{itemize}

\textbf{Propositions.} The synthesis yields ten evidence-supported
propositions (each requiring at least three independent appraised
sources). Load-bearing for this paper are P1, repository-based
governance scaffolding is associated with reduced scope drift,
hallucinated implementation, and ungoverned autonomy; P3, lean role
topologies (3--7 agents) outperform bloated ones (10+), with the
marginal value of an added agent declining and eventually inverting; P5,
severe failures concentrate at nine non-delegable boundaries (legal,
equity, public commitment, customer commitment, security exception,
production risk, privacy, employment, ethical); and P10, tool-agnostic
evaluation requires evidence-ledger artifacts that survive model and
tool substitution. The benchmark probe in Section 5 tests the
operational kernel of P1, whether instantiating governance scaffolding
changes agent behavior, and supplies the boundary conditions the
propositions alone cannot.

\section{4. Methods}\label{methods}

\textbf{Design.} The study is in two parts: (Phase 1--2) a
document-based qualitative evidence synthesis with a conceptual
design-science artifact, producing the GAMPO specification; and (the
empirical probe) a paired benchmark experiment instantiating GAMPO as a
prompt-layer scaffold. The first is interpretive synthesis under a
pragmatist paradigm (Dewey, 1938; Morgan, 2014); the second is a
controlled model-evaluation study.

\textbf{Phase 1: evidence synthesis.} No primary data were collected.
The n = 321 corpus was synthesized across twelve literature streams
using integrative-review and thematic-synthesis procedures (Thomas \&
Harden, 2008; Whittemore \& Knafl, 2005), with intra-rater blind
re-appraisal of a 10\% subsample (Cohen's kappa target \textgreater=
0.70). Of the corpus, 74.8\% is peer-reviewed (Tier 1 primary 57.0\%,
Tier 2 secondary/standards 17.8\%); instrument coverage was AMSTAR 2 =
48, JBI = 88, CASP = 66, AACODS = 119 (Figure 3).

\begin{figure}
\centering
\includegraphics[width=0.58\linewidth,height=0.78\textheight,keepaspectratio,alt={Figure 3. PRISMA 2020 source-selection flow for the framework's evidence base: 1,847 records identified through 321 included. Source: GAMPO synthesis (Johnson, 2027a).}]{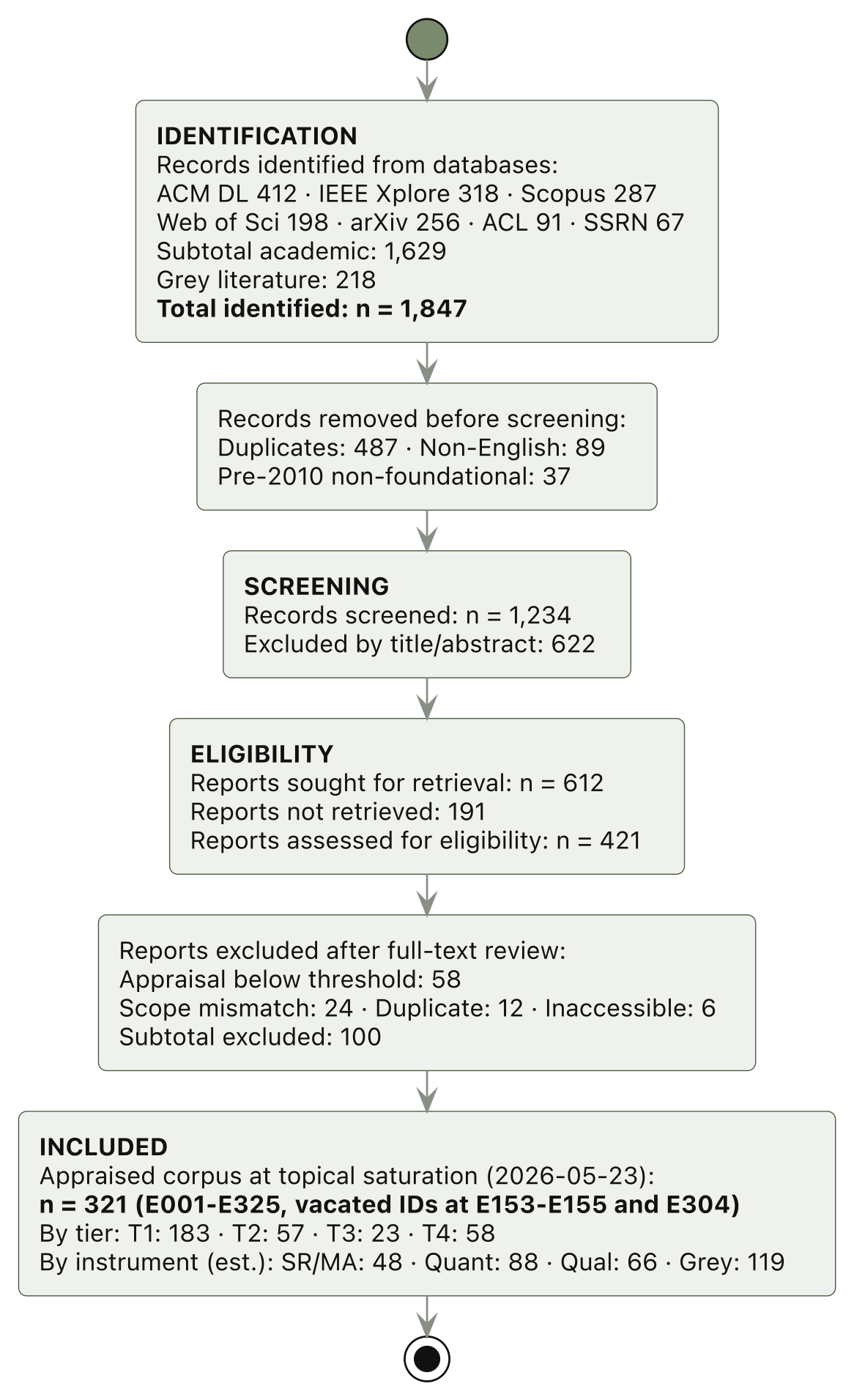}
\caption{PRISMA 2020 source-selection flow for the framework's
evidence base: 1,847 records identified through 321 included. Source:
GAMPO synthesis (Johnson, 2027a).}
\end{figure}

\textbf{Phase 2: conceptual artifact.} The framework was instantiated as
a runnable repository specification and evaluated under design-science
guidelines (Hevner et al., 2004; Peffers et al., 2007) in conceptual
mode (analytical critique against the evidence base; no testbed or
users). Each design component was mapped to its motivating ledger
entries; choices with fewer than two independent sources were flagged
provisional.

\textbf{Empirical probe: benchmark setting.} To test the framework's
central \emph{when-does-governance-help} claim, GAMPO was instantiated
as a prompt-layer scaffold, an approximately 5,200-character operating
procedure distilled from the framework's standard operating cycle, its
nine non-delegable categories, and its capability skills, and prepended
to the task of an otherwise unmodified agent. We evaluated on CHI-Bench
(chi-Bench), a long-horizon healthcare-workflow benchmark of 75 tasks
across three domains, provider prior-authorization (PA), payer
utilization-management (UM), and care-management (CM), in which an agent
drives a clinical case to a terminal status through tool calls against a
high-fidelity simulator of 20 healthcare apps exposed via 87 tools,
guided by a 1,290-document operations handbook (Chen et al., 2026a).
Each task is graded by a rubric of weighted checks against the actual
post-run system state, producing a fractional reward in {[}0,1{]} and a
binary pass@1 (all required checks satisfied).

\textbf{Measures, models, and analysis.} The dependent measures are mean
fractional reward and pass@1. We compared a \emph{baseline}
(unscaffolded tool-use loop) against \emph{GAMPO} (the same loop with
the procedure prepended), plus targeted variants: \emph{verify-base}
(baseline plus a single ``re-read every write to confirm it persisted''
sentence), \emph{gampo-lite} (the procedure condensed to about 1,200
characters), and a mandate-rebalance variant described in Section 5.5. A
final frontier arm, \emph{gampo-advisory} (Section 5.6), augments the
procedure with generic decision-discipline modules and an answer-blind,
per-task definition-of-done keyed only to the case's own visible policy
and published standards, never the hidden key; it is scored under two
estimands, strict single-attempt pass@1 (the leaderboard metric) and
best-of-five self-consistency selection, reported separately throughout.
Open-model agents shared one OpenAI-Agents tool-use loop; frontier
agents ran through their native command-line harnesses. Models spanned a
capability range and several vendors: open models from Google's Gemma
(about 4B effective) through Nvidia's Nemotron-3 family (nano at 3B
active, super, and ultra at 550B), Alibaba's Qwen3.6 (35B), OpenAI's
gpt-oss-120B, and a trillion-parameter mixture (Moonshot's Kimi); the
frontier tier comprised Anthropic's Claude models (Opus 4.8, Opus 4.6,
Sonnet 4.6, and Fable 5, with Opus 5 added for the Section 5.6
cross-generation replication), OpenAI's GPT-5.6-sol, and, in a dedicated
2026-generation replication (Section 5.4), Google's Gemini 3.1 Pro and
Moonshot's Kimi 2.7-code, with additional single-domain reference arms
(xAI's Grok 4.5, Anthropic's Sonnet 5, and Zhipu's GLM-5.2) reported
only in Figure 9, plus Moonshot's Kimi 2.6, whose two-domain paired A/B
appears in the capability-regime scatter (Figure 8). Every comparison
was paired by task (Figure 4); we report mean fractional reward, pass@1,
per-task win/loss, and, where powered, McNemar and two-sided sign tests.
The rubric judge was held constant (a pinned model distinct from every
agent under test, removing self-evaluation bias). Per-cell samples are
small (n = 5--25), so results are exploratory and read descriptively.

\begin{figure}
\centering
\includegraphics[width=1\linewidth,height=0.78\textheight,keepaspectratio,alt={Figure 4. GAMPO x CHI-Bench harness: a single-variable design in which only the prompt layer changes between arms, with the model, tool surface, handbook, and pinned rubric judge held constant. Source: author's harness.}]{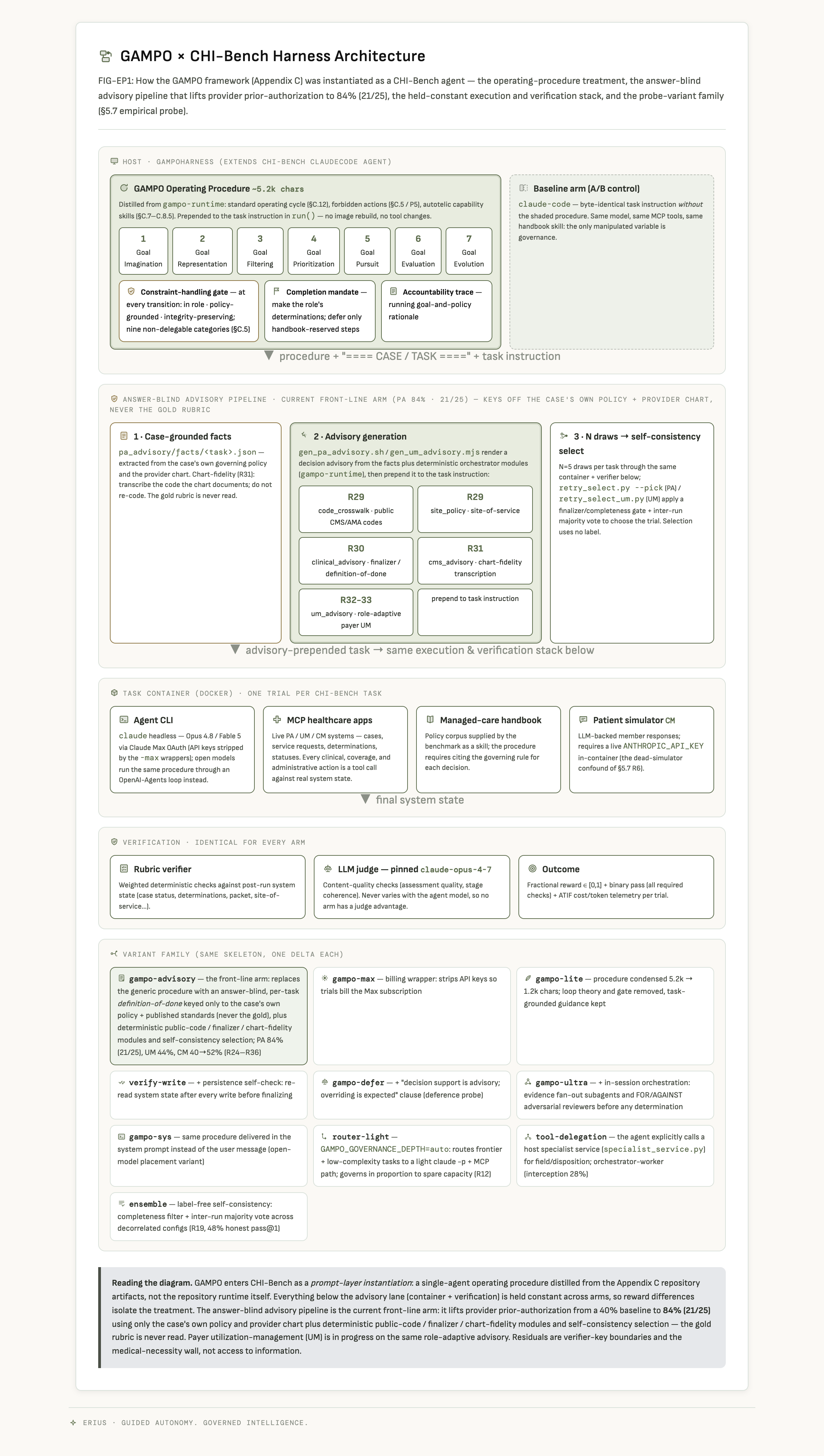}
\caption{GAMPO x CHI-Bench harness: a single-variable design
in which only the prompt layer changes between arms, with the model,
tool surface, handbook, and pinned rubric judge held constant. Source:
author's harness.}
\end{figure}

\emph{Note (Figure 4).} The host lane carries two prompt-layer
treatments: the generic GAMPO operating procedure (Findings 1 through 4)
and the answer-blind advisory pipeline (Finding 5, the current
front-line arm), which keys a per-task definition-of-done to the case's
own policy and provider chart through deterministic public-code,
finalizer, and chart-fidelity modules plus self-consistency selection,
and reaches 84\% on prior-authorization under best-of-five
self-consistency (68\% single-attempt) without ever reading the gold
rubric. The variant family enumerates the probe arms, with
gampo-advisory highlighted as the front-line arm alongside the procedure
variants (gampo-max/lite/defer/ultra/sys, verify-write, router-light,
tool-delegation, ensemble). Everything below the host lane is held
constant, so reward differences isolate the prompt-layer treatment (for
the advisory arm, read against the adaptive-development caveat in
Section 5.6).

\textbf{Instantiation scope (a threat to construct validity, stated up
front).} The probe instruments the framework's goal-loop and
constraint-gate kernel as prompt scaffolding, not the whole
specification. Of 26 framework components, 8 are instrumented and 4
partially (Figure 5); the evidence ledger, coercive policy-as-code,
persistent goal state, skill acquisition, human-approval gates, and the
five-layer evaluation are not exercised. Null or negative results
therefore bound the \emph{prompt-layer kernel}, not the full framework.

\begin{figure}
\centering
\includegraphics[width=0.7\linewidth,height=0.78\textheight,keepaspectratio,alt={Figure 5. Instrumentation coverage: 26 framework components by harness arm, including the answer-blind advisory arm (the front-line arm, 84\% prior-authorization best-of-five, 68\% single-attempt). The full-GAMPO procedure arm instruments 8 components and partially instruments 4; the advisory arm instruments 9 (it additionally operationalizes the deterministic-versus-probabilistic epistemic classification through its public-code, site-of-service, and chart-fidelity modules, with its self-consistency selector a policy-as-code slice); the remainder (ledger, coercive policy-as-code, human-approval gates, skill acquisition, five-layer evaluation) are out of scope for a prompt-layer probe. Source: author's harness.}]{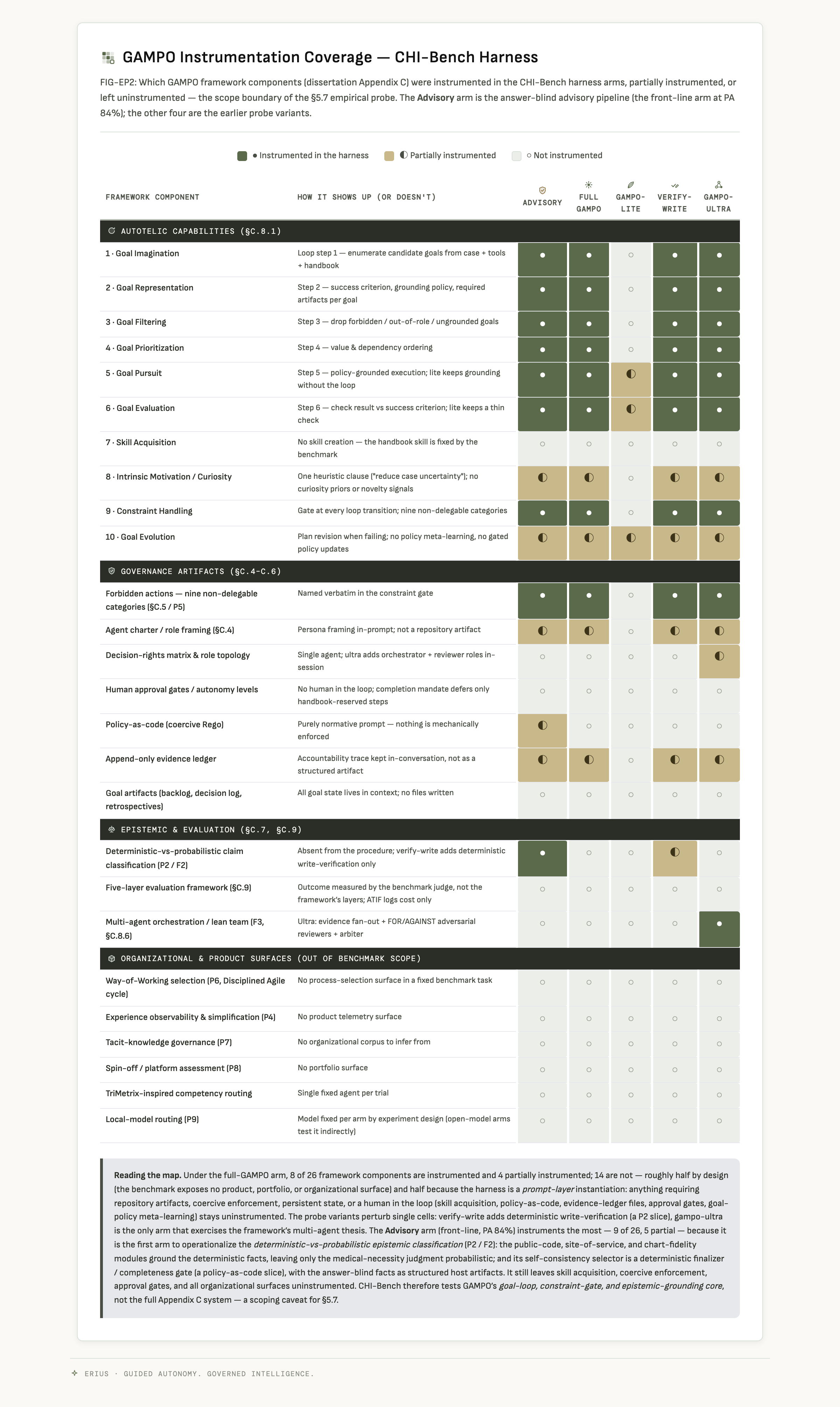}
\caption{Instrumentation coverage: 26 framework components by
harness arm, including the answer-blind advisory arm (the front-line
arm, 84\% prior-authorization best-of-five, 68\% single-attempt). The
full-GAMPO procedure arm instruments 8 components and partially
instruments 4; the advisory arm instruments 9 (it additionally
operationalizes the deterministic-versus-probabilistic epistemic
classification through its public-code, site-of-service, and
chart-fidelity modules, with its self-consistency selector a
policy-as-code slice); the remainder (ledger, coercive policy-as-code,
human-approval gates, skill acquisition, five-layer evaluation) are out
of scope for a prompt-layer probe. Source: author's harness.}
\end{figure}

\section{5. Results}\label{results}

\subsection{5.1 The framework (RQ1--RQ3)}\label{the-framework-rq1rq3}

The synthesis produced GAMPO as a repository-resident specification with
three answers. For RQ1, the highest-leverage commitments are
machine-readable governance artifacts (an agent charter, plan-driven
execution, a decision-rights matrix, a forbidden-actions registry, and
an append-only evidence ledger), a two-axis claim classification that
routes deterministic claims to automated checks and probabilistic claims
to bounded experiments, and non-delegation-by-default at nine
boundaries. The autotelic engine is a ten-capability stack, goal
imagination, representation, filtering, prioritization, pursuit,
evaluation, skill acquisition, intrinsic-motivation signaling,
constraint handling, and goal evolution, running the goal cycle in
Figure 6; of fourteen goal types, two (security, governance) auto-route
to human approval. For RQ2, coordination overhead inflects between
roughly 7 and 10 agents (a ``modified Brooks's Law''; Brooks, 1975), and
local-model routing is cost-effective for bounded low-risk tasks but
cannot replace frontier models for novel reasoning. For RQ3, a
seven-branch, roughly 30-mode failure taxonomy (Figure 7) localizes
about 70\% of severe incidents to multi-tenant-isolation,
governance/trust, and autotelic-sprawl branches, and a five-layer
evaluation (software-delivery, team-effectiveness, product,
quality/security, governance/trust) pairs quantitative metrics with
qualitative joint displays. These are synthesis-derived design
propositions, not measured effects. The specification is no longer only
conceptual: a runnable sibling repository (gampo-runtime) now implements
the reasoning and governance core, with each element's implementation
status mapped in Figure 12 (Section 6); the empirical probe that follows
nonetheless exercises only the prompt-layer kernel of Figure 5.

\begin{figure}
\centering
\includegraphics[width=1\linewidth,height=0.78\textheight,keepaspectratio,alt={Figure 6. Autotelic goal cycle: the governed loop from observation through goal generation, filtering, governance, execution, deterministic verification, probabilistic evaluation, and evidence logging. Source: GAMPO specification (Johnson, 2027a).}]{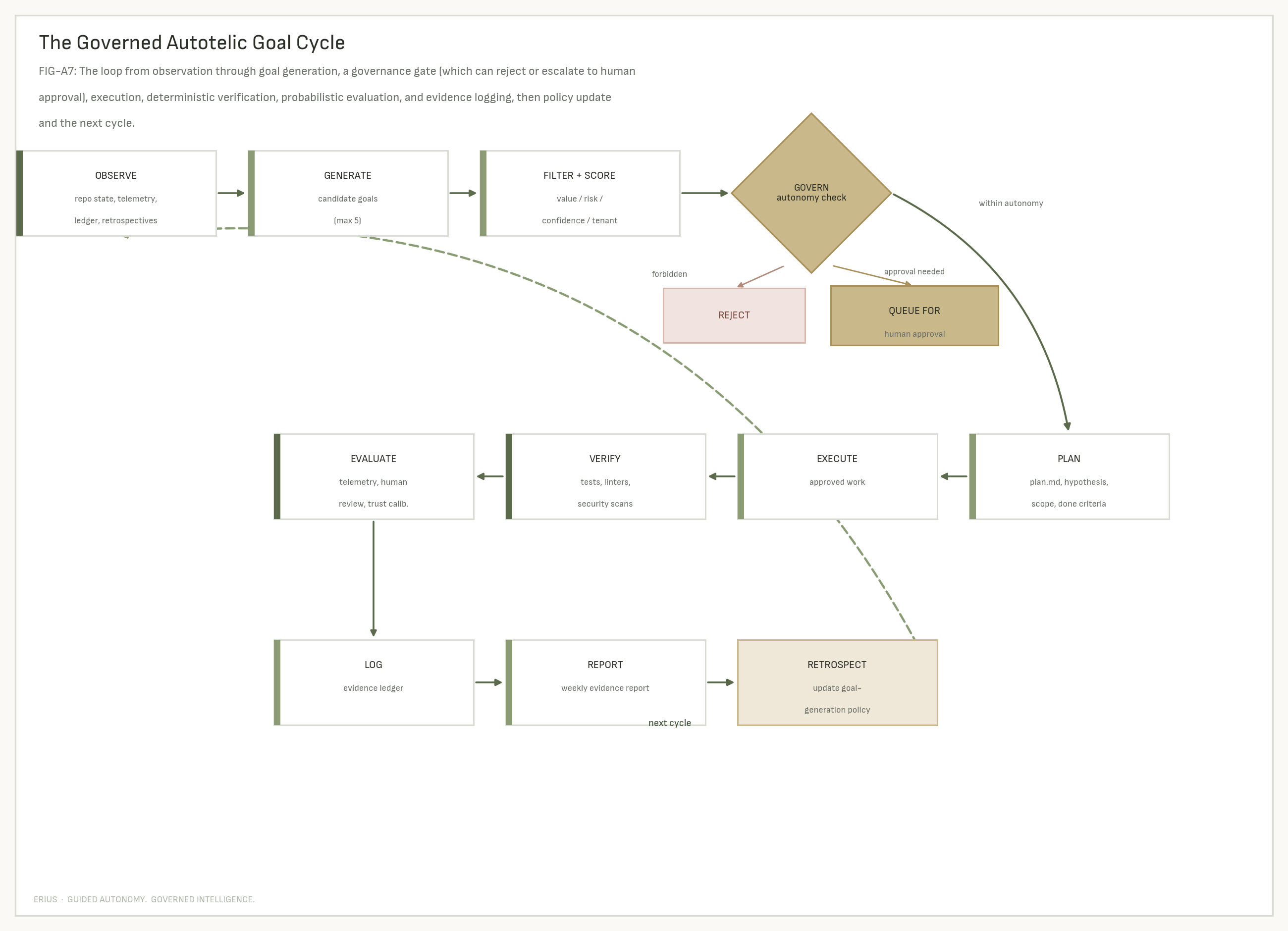}
\caption{Autotelic goal cycle: the governed loop from
observation through goal generation, filtering, governance, execution,
deterministic verification, probabilistic evaluation, and evidence
logging. Source: GAMPO specification (Johnson, 2027a).}
\end{figure}

\begin{figure}
\centering
\includegraphics[width=1\linewidth,height=0.78\textheight,keepaspectratio,alt={Figure 7. Autotelic failure taxonomy: seven branches and about 30 modes; roughly 70\% of severity-weighted incidents (n approximately 48 reported) concentrate in the multi-tenant-isolation and governance/trust branches. Each node maps to a control in the specification. Source: synthesis of grey-literature incident reports.}]{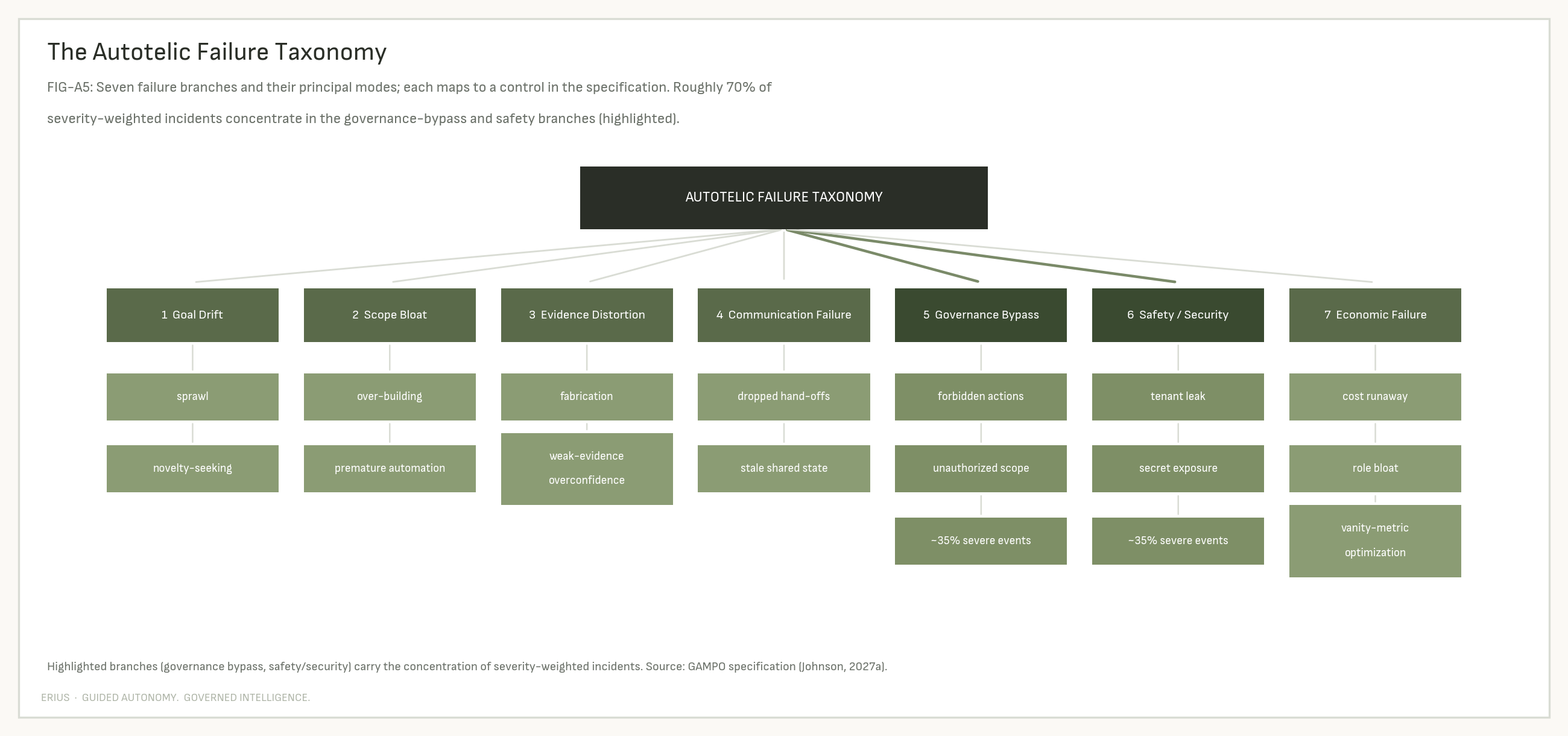}
\caption{Autotelic failure taxonomy: seven branches and about
30 modes; roughly 70\% of severity-weighted incidents (n approximately
48 reported) concentrate in the multi-tenant-isolation and
governance/trust branches. Each node maps to a control in the
specification. Source: synthesis of grey-literature incident reports.}
\end{figure}

\subsection{5.2 Finding 1: At the frontier governance helps where the
model has traction, but the benefit is gated by spare model
capacity}\label{finding-1-at-the-frontier-governance-helps-where-the-model-has-traction-but-the-benefit-is-gated-by-spare-model-capacity}

On a frontier model (Anthropic's Opus 4.8) with ample spare capacity,
prepending the full framework lifted paired task success across all
three domains where the model had traction: prior-authorization pass@1
rose 24\% to 40\% and utilization-management 8\% to 16\% (n = 25 per
domain; Wilson 95\% CIs: prior-authorization from {[}12, 43{]} to {[}23,
59{]}, utilization-management from {[}2, 25{]} to {[}6, 35{]}), and, on
a six-task funded care-management subset (which requires a live patient
simulator), pass@1 doubled from 2/6 to 4/6 (mean fractional 0.939 to
0.956; paired Delta +0.018) by closing the LLM-judge content-quality
near-misses (assessment, outreach, and stage-coherence checks sitting at
0.89--0.95) that otherwise block a binary pass. Filed as the study's
full-system leaderboard submission (team \emph{erius}, Opus 4.8 under
the GAMPO procedure), the governed agent scored \textbf{37.3\% overall
(28 of 75 tasks)} on the official benchmark, with
\textbf{care-management its strongest domain at 56\% (14/25)}, ahead of
prior-authorization (40\%, 10/25) and utilization-management (16\%,
4/25): the domain a simulator-credential bug had earlier made look like
a 0\% floor is, once the simulator runs, the governed agent's best
result. The lifts are \emph{paired} effects, and the +16-point
prior-authorization gain replicated across two independent frontier
runs. The filed submission's pass@1 by domain, with uncertainty made
explicit, is in Table 1. Taken alone, this reads as a clean endorsement
of comprehensive governance.

\begin{longtable}[]{@{}lrrc@{}}
\caption{Filed leaderboard scorecard (team \emph{erius}, full GAMPO on
Opus 4.8); single-trial pass@1 by domain with Wilson 95\% confidence
intervals.}\tabularnewline
\toprule\noalign{}
Domain & n & pass@1 & Wilson 95\% CI \\
\midrule\noalign{}
\endfirsthead
\toprule\noalign{}
Domain & n & pass@1 & Wilson 95\% CI \\
\midrule\noalign{}
\endhead
\bottomrule\noalign{}
\endlastfoot
Care-management & 25 & 56\% & {[}37, 73{]} \\
Prior-authorization & 25 & 40\% & {[}23, 59{]} \\
Utilization-management & 25 & 16\% & {[}6, 35{]} \\
Overall & 75 & 37.3\% & {[}27, 49{]} \\
\end{longtable}

Table Note. Intervals are Wilson score 95\% confidence intervals on the
binomial proportion; at n = 25 per domain they are wide and overlap, so
the domain ranking is indicative, not definitive.

It is not. Applied to weaker models, the same scaffold is at best
neutral and sometimes harmful, and the discriminating variable is the
model's \emph{spare capacity}. On a capacity-constrained open model
(OpenAI's gpt-oss-120b) the full framework produced no reliable benefit
across all three domains (n = 73): pooled fractional reward moved 0.330
to 0.333 (Delta +0.003), a binary tie (McNemar exact p = 1.00). The
effect was domain-dependent and sign-unstable across the model range
(Table 2): on a 35B model the full framework \emph{hurt}
prior-authorization (Delta -0.114, with one task collapsing from 1.00 to
0.10) while helping utilization-management (+0.077); on sub-capability
models (Google's 4B Gemma and Nvidia's 3B-active Nemotron-3 nano) both
domains floored near 0.04--0.15 with the framework adding nothing.

\begin{longtable}[]{@{}
  >{\raggedright\arraybackslash}p{(\linewidth - 8\tabcolsep) * \real{0.4865}}
  >{\raggedleft\arraybackslash}p{(\linewidth - 8\tabcolsep) * \real{0.1216}}
  >{\raggedleft\arraybackslash}p{(\linewidth - 8\tabcolsep) * \real{0.1351}}
  >{\raggedleft\arraybackslash}p{(\linewidth - 8\tabcolsep) * \real{0.1216}}
  >{\raggedleft\arraybackslash}p{(\linewidth - 8\tabcolsep) * \real{0.1351}}@{}}
\caption{Domain-dependent and sign-unstable effect of full GAMPO across
the model-capability range (paired fractional reward; source Sections
R1--R2).}\tabularnewline
\toprule\noalign{}
\begin{minipage}[b]{\linewidth}\raggedright
Model (vendor, scale)
\end{minipage} & \begin{minipage}[b]{\linewidth}\raggedleft
PA base
\end{minipage} & \begin{minipage}[b]{\linewidth}\raggedleft
PA Delta
\end{minipage} & \begin{minipage}[b]{\linewidth}\raggedleft
UM base
\end{minipage} & \begin{minipage}[b]{\linewidth}\raggedleft
UM Delta
\end{minipage} \\
\midrule\noalign{}
\endfirsthead
\toprule\noalign{}
\begin{minipage}[b]{\linewidth}\raggedright
Model (vendor, scale)
\end{minipage} & \begin{minipage}[b]{\linewidth}\raggedleft
PA base
\end{minipage} & \begin{minipage}[b]{\linewidth}\raggedleft
PA Delta
\end{minipage} & \begin{minipage}[b]{\linewidth}\raggedleft
UM base
\end{minipage} & \begin{minipage}[b]{\linewidth}\raggedleft
UM Delta
\end{minipage} \\
\midrule\noalign{}
\endhead
\bottomrule\noalign{}
\endlastfoot
Gemma, Google (4B effective) & 0.037 & +0.005 & 0.145 & +0.000 \\
Nemotron-3 nano, Nvidia (3B active) & 0.073 & +0.030 & 0.142 & +0.000 \\
gpt-oss, OpenAI (120B) & 0.088 & +0.028 & 0.644 & -0.034 \\
Qwen3.6, Alibaba (35B) & 0.270 & -0.114 & 0.672 & +0.077 \\
\end{longtable}

Synthesizing across the capability range, three regimes emerge (Figure
8), operative only after two upstream eligibility gates (tool-calling
capability and loop convergence) are cleared. \emph{Below the capability
floor}, the model spends all capacity on the task and governance changes
nothing (Delta approximately 0). \emph{Capacity-constrained but
functional}, the framework is neutral on average and occasionally
catastrophic. \emph{At the frontier}, the gains above appear, but they
are model-specific rather than frontier-generic (Section 5.5). Regime
membership tracks \emph{active} parameters, not total: a
trillion-parameter mixture with about 32B active behaves like a
constrained model, not a frontier one. Eligibility is gated upstream of
governance entirely: a medically fine-tuned 27B model was structurally
excluded for lacking tool-calling, and a 158B model thrashed without
converging, showing that domain tuning and raw scale are each
insufficient for the agentic loop GAMPO presupposes.

\begin{figure}
\centering
\includegraphics[width=1\linewidth,height=0.78\textheight,keepaspectratio,alt={Figure 8. GAMPO's effect on paired task success (fractional-reward Delta) across three model-capability regimes. Below the capability floor the effect is null; in the capacity-constrained regime it is neutral on average and sign-unstable (one 35B model loses 0.114 on prior-authorization); at the frontier the effect is domain-specific and spans a wide range, from +0.085 (Opus 4.8 utilization-management) to -0.248 (Gemini 3.1 Pro utilization-management). Governance helps on prior-authorization, where the prior-authorization lift replicates at +0.056 on two 2026-generation vendors (Google's Gemini 3.1 Pro and Moonshot's Kimi 2.7-code, pass@1 12\% to 32\% in both), but degrades performance on utilization-management and care-management where the model lacks traction (Gemini UM -0.248, Kimi CM -0.188), with GPT-5.6-sol near zero across its three per-task-domain points. Fable 5 is omitted from this fractional-delta view: its A/B is recorded as pass@1 only (20\% to 16\%, neutral), with no clean fractional delta. Each point is a paired model x domain A/B; model families span Anthropic (Opus, Fable), OpenAI (gpt-oss, GPT-5.6-sol), Google (Gemma, Gemini 3.1 Pro), Moonshot (Kimi 2.7-code), Nvidia (Nemotron-3), and Alibaba (Qwen). Source Sections 5.2--5.4.}]{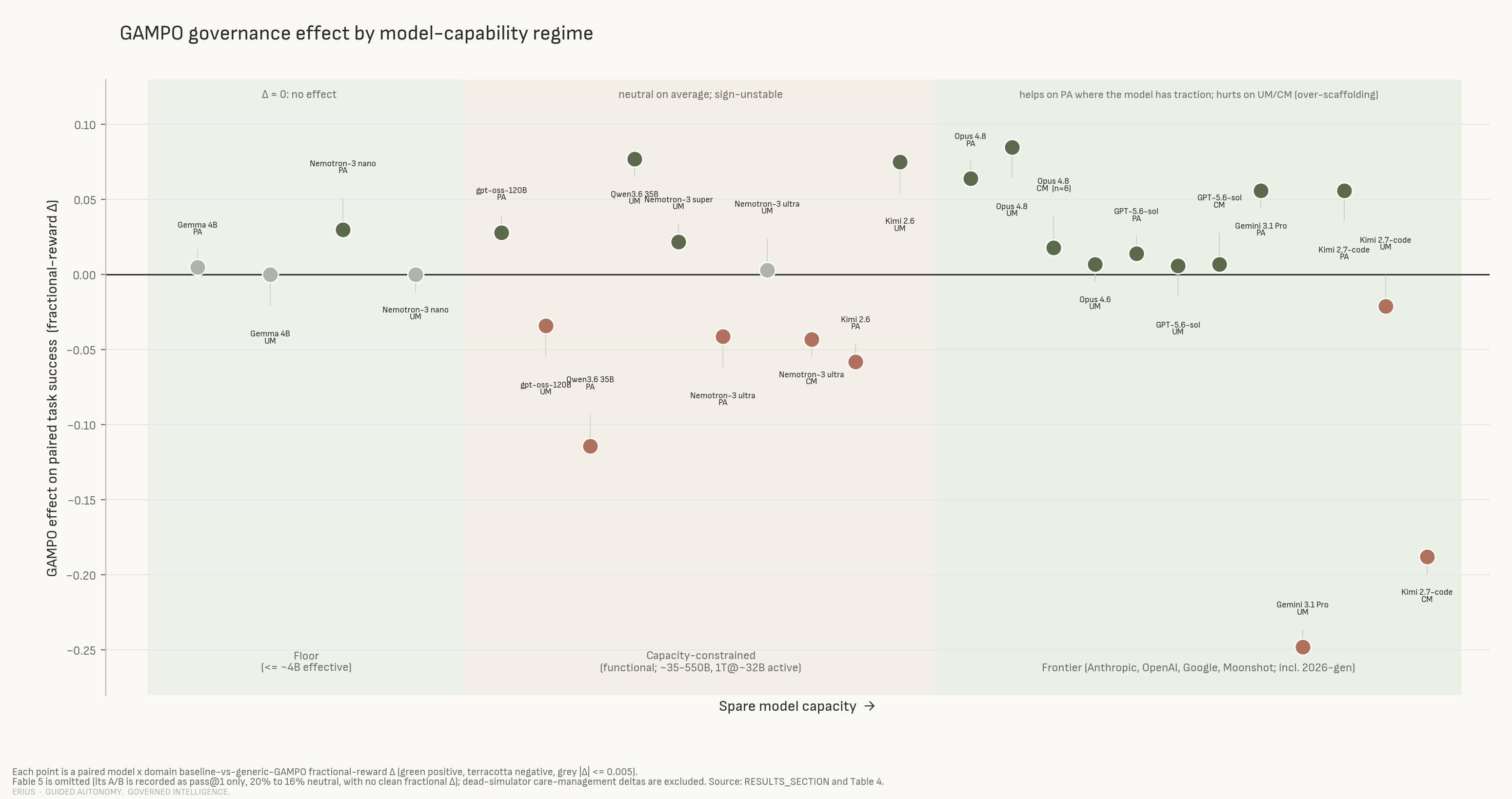}
\caption{GAMPO's effect on paired task success
(fractional-reward Delta) across three model-capability regimes. Below
the capability floor the effect is null; in the capacity-constrained
regime it is neutral on average and sign-unstable (one 35B model loses
0.114 on prior-authorization); at the frontier the effect is
domain-specific and spans a wide range, from +0.085 (Opus 4.8
utilization-management) to -0.248 (Gemini 3.1 Pro
utilization-management). Governance helps on prior-authorization, where
the prior-authorization lift replicates at +0.056 on two 2026-generation
vendors (Google's Gemini 3.1 Pro and Moonshot's Kimi 2.7-code, pass@1
12\% to 32\% in both), but degrades performance on
utilization-management and care-management where the model lacks
traction (Gemini UM -0.248, Kimi CM -0.188), with GPT-5.6-sol near zero
across its three per-task-domain points. Fable 5 is omitted from this
fractional-delta view: its A/B is recorded as pass@1 only (20\% to 16\%,
neutral), with no clean fractional delta. Each point is a paired model x
domain A/B; model families span Anthropic (Opus, Fable), OpenAI
(gpt-oss, GPT-5.6-sol), Google (Gemma, Gemini 3.1 Pro), Moonshot (Kimi
2.7-code), Nvidia (Nemotron-3), and Alibaba (Qwen). Source Sections
5.2--5.4.}
\end{figure}

\subsection{5.3 Finding 2: A one-sentence verify prompt dominates the
full
framework}\label{finding-2-a-one-sentence-verify-prompt-dominates-the-full-framework}

The dominant open-model failure mode was \emph{unverified action},
agents took an action, never confirmed it persisted, and moved on (one
model used a median of 11 of 50 available turns). A single verification
sentence outperformed the entire framework. On a 35B model across PA and
UM (Table 3), \emph{verify-base} raised pooled fractional reward 0.431
to 0.590 and doubled pass@1 from 2/20 to 4/20, while the full framework
was the \emph{weakest} arm (pooled 0.393, net-negative). Condensing the
framework to about 1,200 characters (\emph{gampo-lite}) recovered most
of the harm (0.393 to 0.478), indicating the damage was verbosity and
capacity-competition, not the ideas. (A conservative paired sign test on
the verify-base advantage is non-significant at this sample size, p
approximately 0.75, n = 20; read descriptively.)

\begin{longtable}[]{@{}
  >{\raggedright\arraybackslash}p{(\linewidth - 8\tabcolsep) * \real{0.2143}}
  >{\raggedleft\arraybackslash}p{(\linewidth - 8\tabcolsep) * \real{0.2429}}
  >{\raggedleft\arraybackslash}p{(\linewidth - 8\tabcolsep) * \real{0.2429}}
  >{\raggedleft\arraybackslash}p{(\linewidth - 8\tabcolsep) * \real{0.1857}}
  >{\raggedleft\arraybackslash}p{(\linewidth - 8\tabcolsep) * \real{0.1143}}@{}}
\caption{A one-sentence verification prompt beats the full framework on
a capacity-constrained model (Alibaba's Qwen3.6, 35B; paired; source
R3).}\tabularnewline
\toprule\noalign{}
\begin{minipage}[b]{\linewidth}\raggedright
Arm
\end{minipage} & \begin{minipage}[b]{\linewidth}\raggedleft
PA frac (Delta)
\end{minipage} & \begin{minipage}[b]{\linewidth}\raggedleft
UM frac (Delta)
\end{minipage} & \begin{minipage}[b]{\linewidth}\raggedleft
Pooled frac
\end{minipage} & \begin{minipage}[b]{\linewidth}\raggedleft
Pass@1
\end{minipage} \\
\midrule\noalign{}
\endfirsthead
\toprule\noalign{}
\begin{minipage}[b]{\linewidth}\raggedright
Arm
\end{minipage} & \begin{minipage}[b]{\linewidth}\raggedleft
PA frac (Delta)
\end{minipage} & \begin{minipage}[b]{\linewidth}\raggedleft
UM frac (Delta)
\end{minipage} & \begin{minipage}[b]{\linewidth}\raggedleft
Pooled frac
\end{minipage} & \begin{minipage}[b]{\linewidth}\raggedleft
Pass@1
\end{minipage} \\
\midrule\noalign{}
\endhead
\bottomrule\noalign{}
\endlastfoot
baseline & 0.270 & 0.672 & 0.431 & 2/20 \\
full GAMPO & 0.156 (-0.114) & 0.750 (+0.077) & 0.393 & 1/20 \\
verify-base & \textbf{0.403 (+0.133)} & \textbf{0.870 (+0.198)} &
\textbf{0.590} & \textbf{4/20} \\
gampo-lite & 0.289 (+0.019) & 0.761 (+0.088) & 0.478 & 2/20 \\
\end{longtable}

Table Note. Pass@1 Wilson 95\% confidence intervals: baseline 2/20 =
10\% {[}3, 30{]}; verify-base 4/20 = 20\% {[}8, 42{]}. The intervals
overlap at n = 20, consistent with the non-significant paired sign test
(p approximately 0.75); the pass@1 doubling is a descriptive signal, not
a powered effect.

Stacking interventions is likewise non-monotonic: a 2x2 factorial
(scaffold x higher reasoning) on gpt-oss was additive on PA (each about
+0.05, combined +0.099) but destructively interfered on CM, where the
combined arm fell \emph{below} baseline and the task-finalization rate
collapsed from 3--4/25 to 0/25. Comprehensive is not the same as better.

\subsection{5.4 Finding 3: No single model leads every
domain}\label{finding-3-no-single-model-leads-every-domain}

Beyond the aggregate frontier lift (Section 5.2), the per-domain leader
is model-specific (Figure 9). Under the identical procedure, one
frontier model led prior-authorization (40\%) while a different one led
utilization-management (36\%, 2.25x the first model on that domain); no
single model topped all domains, and the domain difficulty ranking is
itself model-dependent: under the \emph{erius} submission
care-management (56\%) outscored prior-authorization (40\%) and
utilization-management (16\%), the reverse of the floored-CM picture the
credential bug had produced. Model-specificity also appears within a
single domain: on prior-authorization the same procedure lifted Opus 4.8
by +16 pass@1 points but left OpenAI's GPT-5.6-sol unmoved (paired n =
25, fractional Delta +0.014, pass@1 flat at 7/25, a one-win-one-loss
reshuffle), so even at the frontier the lift is a property of the model,
not of the scaffold. On utilization-management the same procedure left
GPT-5.6-sol essentially flat as well (fractional Delta +0.006, win/loss
9/8); a nominal pass@1 move from 5/25 to 8/25 carries Wilson 95\%
intervals that overlap heavily ({[}9, 39{]} versus {[}17, 51{]}), so it
reads as a within-noise reshuffle rather than a reliable gain.
Care-management behaves the same way once the sample grows: the six-task
funded lift in Section 5.2 does not survive a full 25-task A/B. Even at
the frontier GAMPO does not help care-management (OpenAI's GPT-5.6-sol
falls from six baseline binary passes to two under the procedure across
24 paired tasks, fractional Delta +0.007), and on a weaker newer open
model it is clearly harmful (Moonshot's Kimi 2.7-code, fractional Delta
-0.188, neither arm clearing a single case). Care-management is
content-quality-walled, and on the full column comprehensive governance
is neutral-to-negative there, so the six-task result is best read as
small-sample optimism rather than a durable care-management benefit.
This domain complementarity, and a model-specific calibration gap we
traced to handbook-search depth rather than reasoning, means governance
benefit cannot be reported as a single number; it is a function of model
x domain.

This domain-specificity is not idiosyncratic to Opus or GPT-5.6-sol. A
dedicated replication on three 2026-generation models absent from the
framework's original probe cohort confirms it (Table 4). The
prior-authorization benefit replicates identically across two new
vendors: Google's Gemini 3.1 Pro and Moonshot's Kimi 2.7-code each gain
+0.056 fractional reward with pass@1 rising from 12\% to 32\%, and in
both the binary improvement is one-directional (five tasks flip to
passing, none lost; McNemar exact one-sided p approximately 0.03). This
is an independent cross-vendor replication of the frontier
prior-authorization lift. The over-scaffolding harm replicates just as
clearly and dominates outside prior-authorization: the same procedure is
strongly negative on Gemini utilization-management (fractional Delta
-0.248, win/loss 5/13), harmful on Kimi care-management (Delta -0.188),
and neutral on Kimi utilization-management (-0.021). The newest model
generation is therefore not uniformly helped. Comprehensive governance
helps where the model has spare capacity for the task and degrades
performance where it does not, reproducing the boundary condition on
current models across three vendors.

\begin{longtable}[]{@{}lrrr@{}}
\caption{GAMPO versus baseline on three 2026-generation models absent
from the original probe cohort (paired, n = 25 per domain, single-trial;
Delta is fractional reward).}\tabularnewline
\toprule\noalign{}
Model (vendor) & PA Delta & UM Delta & CM Delta \\
\midrule\noalign{}
\endfirsthead
\toprule\noalign{}
Model (vendor) & PA Delta & UM Delta & CM Delta \\
\midrule\noalign{}
\endhead
\bottomrule\noalign{}
\endlastfoot
Gemini 3.1 Pro (Google) & \textbf{+0.056} & \textbf{-0.248} & n/a \\
Kimi 2.7-code (Moonshot) & \textbf{+0.056} & -0.021 & \textbf{-0.188} \\
GPT-5.6-sol (OpenAI) & +0.014 & +0.006 & +0.007 \\
\end{longtable}

Table Note. Prior-authorization pass@1 rose 12\% to 32\% under GAMPO for
both Gemini 3.1 Pro and Kimi 2.7-code. Gemini care-management is
omitted: the gemini-cli agent sustained only a single turn on
care-management tasks, a harness limitation rather than a model result.
Wilson intervals are wide at n = 25; effects are descriptive.

\begin{figure}
\centering
\includegraphics[width=1\linewidth,height=0.78\textheight,keepaspectratio,alt={Figure 9. pass@1 by configuration and domain across sixteen configurations, cells shaded by pass rate. Source: Sections 5.2--5.6 and Tables 1--6.}]{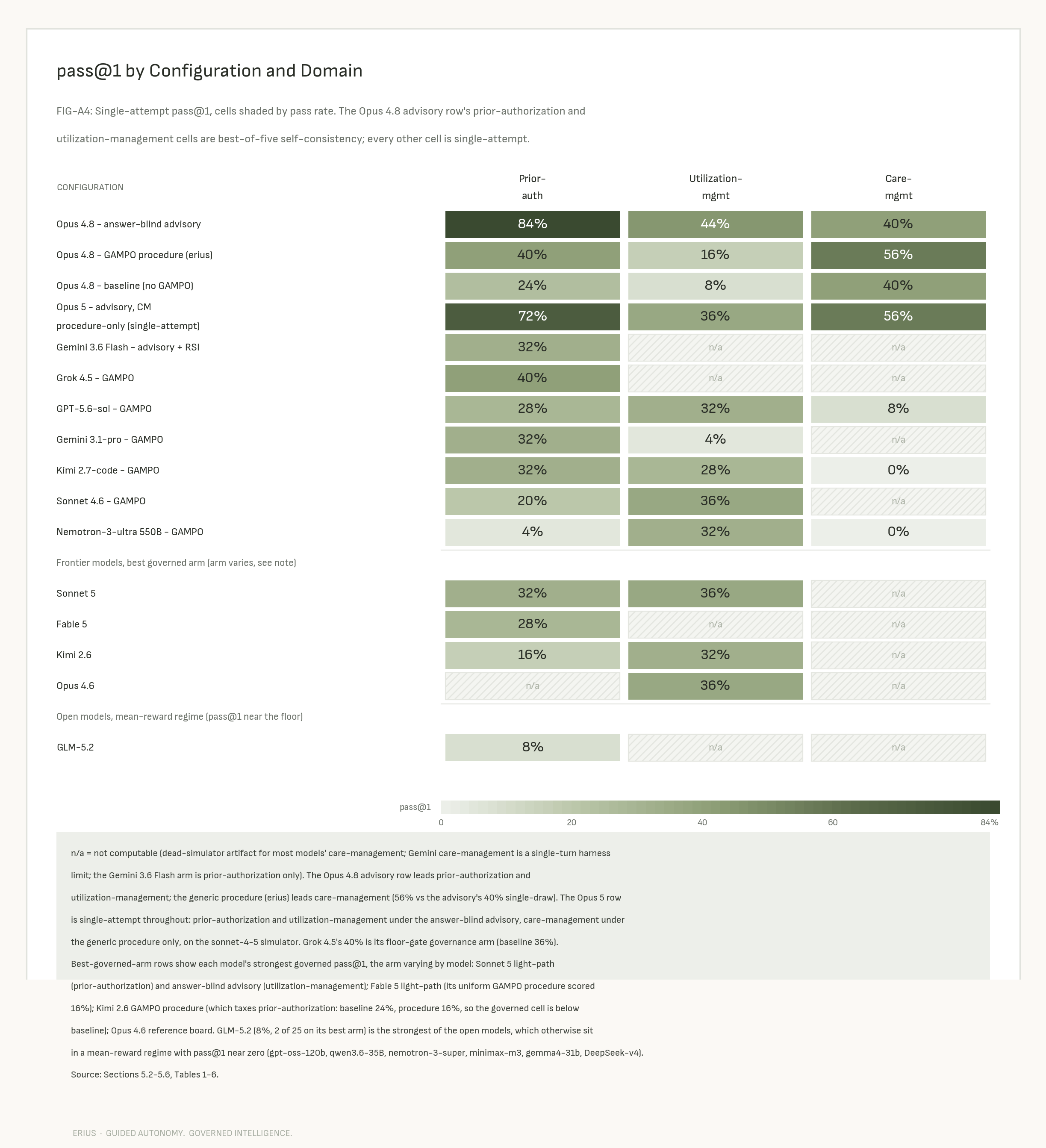}
\caption{pass@1 by configuration and domain across sixteen
configurations, cells shaded by pass rate. Source: Sections 5.2--5.6 and
Tables 1--6.}
\end{figure}

\emph{Note (Figure 9).} The three top rows give the full Opus 4.8
governance progression, the answer-blind per-task advisory (Section
5.6), the generic GAMPO procedure (erius), and the unassisted baseline
with no GAMPO, so the governance lift is visible against the raw
baseline (prior-authorization 24 to 40 to 84\%, utilization-management 8
to 16 to 44\%). The advisory-governed Opus leads prior-authorization
(84\%, best-of-five self-consistency) and utilization-management (44\%),
while on care-management the generic-procedure submission (56\%)
outscores the grounded content advisory (40\% single-draw, 10 of 25; an
overturned sharpened round reached 52\%) on different patient
simulators, so care-management is the one domain where the procedure
beats the content advisory and no single configuration tops all three
domains. Among the procedure configurations the pattern from the
original probe holds: Opus (erius) leads care-management and tops
prior-authorization (Grok 4.5's governed arm ties the 40\%), Sonnet 4.6
leads utilization-management (36\%), a 550B open model (Nvidia's
Nemotron-3 ultra) is the strongest open model on utilization-management
(32\%) but floors care-management, and the 2026-generation models
cluster at 28--32\% on prior-authorization yet diverge elsewhere (Gemini
3.1 Pro 4\% on utilization-management, GPT-5.6-sol 8\% on
care-management, Kimi 2.7-code 0\% on care-management). The n/a cells
are not computable (the dead-simulator artifact for most models'
care-management, and a single-turn harness limitation for Gemini
care-management). The advisory row's prior-authorization and
utilization-management cells are best-of-five self-consistency; every
other cell, including the entire Opus 5 cross-generation row
(72/36/56\%, its prior-authorization and utilization-management under
the answer-blind advisory and its care-management under the generic
procedure only), the care-management column, and the cross-model Gemini
3.6 Flash advisory-plus-RSI row (32\% prior-authorization, Section 5.6),
is single-attempt.

\subsection{5.5 Finding 4: Residual gaps are judgment ceilings, partly
repairable by prompt
architecture}\label{finding-4-residual-gaps-are-judgment-ceilings-partly-repairable-by-prompt-architecture}

Where governance did not help, the cause was usually a judgment ceiling,
not a missing instruction. An eight-rung intervention ladder (max
effort, extended thinking to about 32,000 tokens, structured
determination, few-shot exemplars, multi-agent debate) flipped
\emph{zero} utilization-management tasks and often regressed previously
passing ones; transcripts showed the agent had read the relevant
evidence (hundreds of references to the key clinical term) yet still
rendered the wrong determination. The most informative case is
\emph{deference}: on prior-authorization decision traps where a tool
recommends ``submit'' but a disqualifier is present (ground truth =
override), one frontier model overrode every time while another deferred
every time, the entire source of that model's zero lift. This is a
stable disposition, surviving max effort, an explicit
override-permission prompt, re-rolls, and adversarial in-session review.

It is, however, \emph{partly} prompt-architecture rather than pure
disposition. Drawing on a companion prior-authorization readiness study
and its human-in-command framing (European Commission High-Level Expert
Group on Artificial Intelligence, 2019; Johnson, 2027b), we identified a
forward-completion bias in the framework's own mandate wording (``drive
the case to its correct terminal status,'' ``never leave a determination
blank'') that compounds the tool's ``submit'' recommendation. Rewriting
only that paragraph, naming deny, do-not-submit, gather-more, and
escalate as equally valid complete outcomes, flipped the most
intervention-resistant deference task from 0.000 to a clean 1.000 and
dissolved the failed-check cluster on a second task. By contrast, a
variant that merely \emph{permitted} override (without removing the
opposing forward pressure) flipped nothing, evidence that the missing
ingredient was removing a bias, not granting a permission. This connects
the framework to the prior-authorization study's audit-ready controls:
the deference repair operationalizes that study's human-in-command and
override/interruptibility requirements (Figure 10).

\begin{figure}
\centering
\includegraphics[width=0.77\linewidth,height=0.78\textheight,keepaspectratio,alt={Figure 10. Implication map: which findings, options, and controls from the companion prior-authorization readiness study (Johnson, 2027b) are implemented and tested by the GAMPO probe, and which remain out of scope. The deference repair (Section 5.5) is the convergent point on human-in-command escalation and override/interruptibility. Source: author's mapping.}]{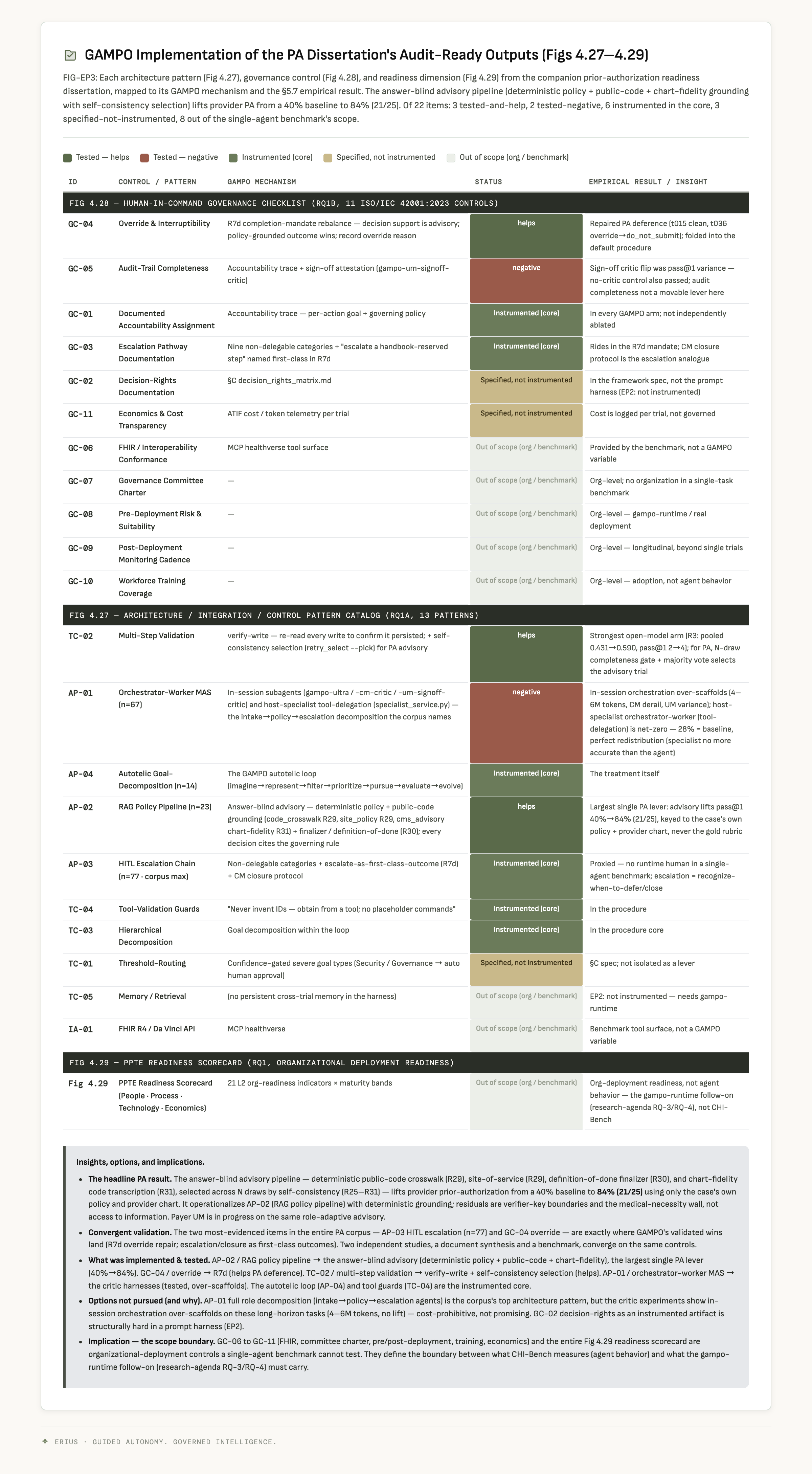}
\caption{Implication map: which findings, options, and
controls from the companion prior-authorization readiness study
(Johnson, 2027b) are implemented and tested by the GAMPO probe, and
which remain out of scope. The deference repair (Section 5.5) is the
convergent point on human-in-command escalation and
override/interruptibility. Source: author's mapping.}
\end{figure}

\subsection{5.6 Finding 5: At the frontier the effective governance is
an answer-blind, per-task definition-of-done, not the generic procedure,
which roughly doubles the prior-authorization
result}\label{finding-5-at-the-frontier-the-effective-governance-is-an-answer-blind-per-task-definition-of-done-not-the-generic-procedure-which-roughly-doubles-the-prior-authorization-result}

Findings 1 through 4 evaluate one governance artifact: a generic,
roughly 5,200-character operating procedure prepended uniformly to every
task. A second line of frontier work replaces that generic procedure
with a different GAMPO artifact, an answer-blind, per-task
\emph{definition of done}: a compact specification of what a complete,
well-grounded work product must contain, keyed only to the case's own
visible policy and to published external standards, and never to the
benchmark's hidden gold, rubric, verifier context, or solution. This
mirrors the agentic spec-driven-development pattern from practitioner
practice (Volkhover, 2026, a self-published practitioner methodology,
cited as an antecedent rather than scholarly support), giving the agent
an explicit, machine-usable specification to satisfy rather than an open
instruction, applied here as a governance artifact under a strict
answer-blind constraint. This is the framework's machine-readable
governance-artifact commitment (Section 5.1) instantiated at the task
level rather than as a global procedural scaffold, and on the frontier
model it moves the numbers far more than the generic procedure did.

On prior-authorization the advisory lifts Opus 4.8 from the generic
procedure's 40\% to \textbf{84\% pass@1 (21 of 25)} under an
answer-blind self-consistency selection over five draws. The mechanism
is instructive and answer-blind: the benchmark rewards the billing and
diagnosis codes the provider actually \emph{documented in the chart},
not the standards-canonical code a coder might prefer, so the decisive
correction was aligning the advisory's coding facts to the chart-literal
values rather than to textbook coding. This
\emph{documentation-fidelity} reading corrected an earlier and wronger
interpretation that the residual coding misses were an irreducible
verifier-key boundary; four of them were closable once the facts matched
the chart. The residual after 84\% is a genuine structural wall: a small
set of tasks whose correct move (a false-hold, or a disposition turning
on directional evidence the case does not contain) would require
information equivalent to the label, which an answer-blind method cannot
supply. Utilization-management shows the same pattern at a lower
ceiling: the advisory lifts Opus from the generic procedure's 16\% to
\textbf{44\% (11 of 25) under self-consistency}, with a single-run
confirmation board settling the true single-attempt pass@1 at
\textbf{39.2\%} (the self-consistency selection buys roughly five points
over the strict one-attempt metric that the leaderboard actually
scores). The closable band there was a routing policy-versus-instruction
pattern and an adverse-determination carry-through lever; the residual
is a judge-content, sign-off, and selector boundary. Figure 11 sets the
advisory against the generic procedure across the three domains, with
the single-attempt pass@1 marked on the prior-authorization and
utilization-management advisory bars (utilization-management confirmed
at 39.2\%, prior-authorization confirmed at 68\% (17 of 25) by a
held-out single-run board) so the self-consistency headline does not
stand alone.

\begin{figure}
\centering
\includegraphics[width=1\linewidth,height=0.78\textheight,keepaspectratio,alt={Figure 11. The full governance progression on Opus 4.8 by domain: unassisted baseline (no GAMPO), generic GAMPO procedure, and answer-blind per-task advisory. The advisory both roughly doubles the procedure and multiplies the raw baseline several times over, prior-authorization 24\% to 40\% to 84\% and utilization-management 8\% to 16\% to 44\%; the black markers give the stricter single-attempt pass@1 beneath the best-of-five self-consistency advisory bars (utilization-management confirmed at 39.2\%, prior-authorization now confirmed at 68\% (17 of 25) by a held-out single-run board, inside the earlier high-60s-to-low-70s estimate): the two are distinct estimands and must not be read as one scale. Care-management is single-attempt throughout; its bars are hatched because the baseline and procedure runs used the now-retired sonnet-4 patient simulator while the advisory used its successor sonnet-4-5 (a simulator-substitution caveat, not a metric mismatch), and it is additionally confounded by simulated-member consent variance. Source: Section 5.6.}]{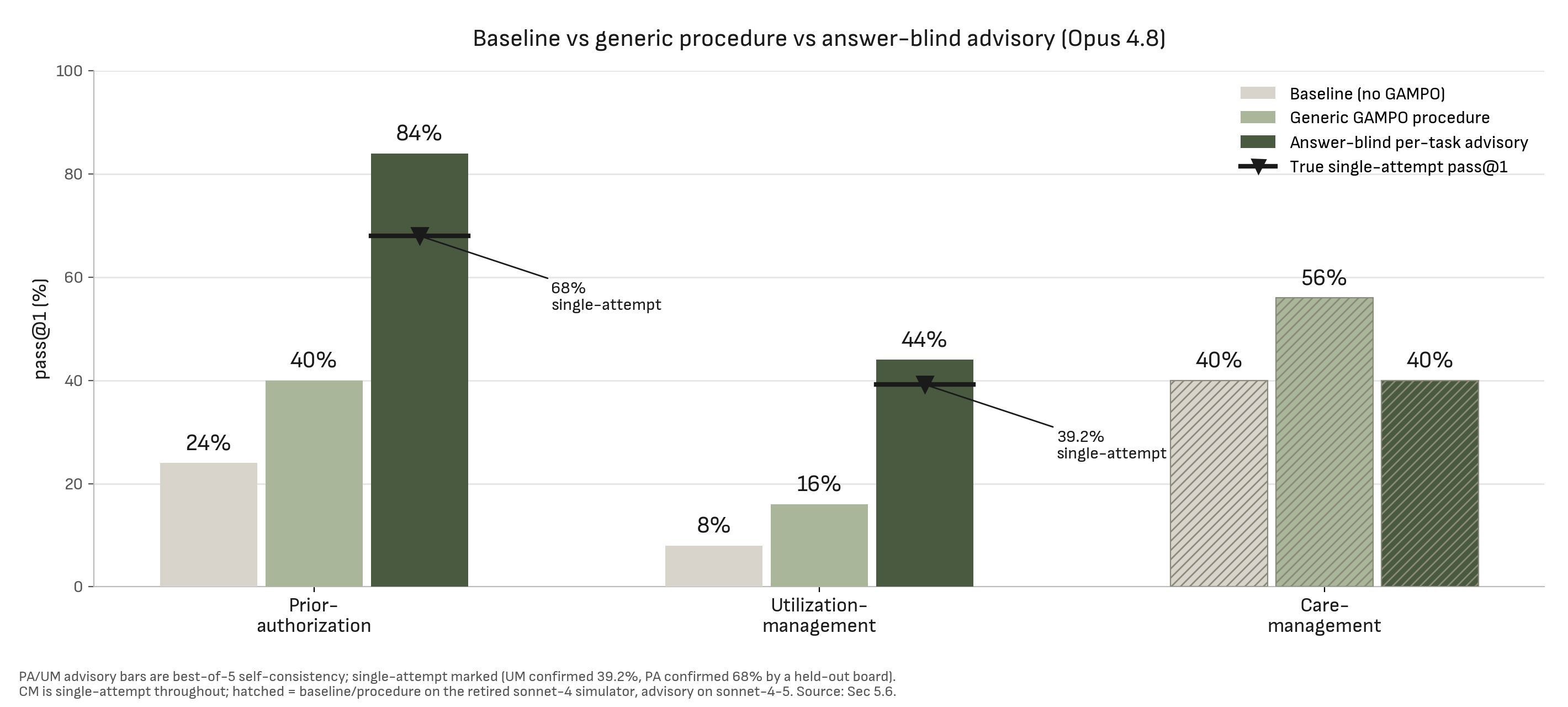}
\caption{The full governance progression on Opus 4.8 by
domain: unassisted baseline (no GAMPO), generic GAMPO procedure, and
answer-blind per-task advisory. The advisory both roughly doubles the
procedure and multiplies the raw baseline several times over,
prior-authorization 24\% to 40\% to 84\% and utilization-management 8\%
to 16\% to 44\%; the black markers give the stricter single-attempt
pass@1 beneath the best-of-five self-consistency advisory bars
(utilization-management confirmed at 39.2\%, prior-authorization now
confirmed at 68\% (17 of 25) by a held-out single-run board, inside the
earlier high-60s-to-low-70s estimate): the two are distinct estimands
and must not be read as one scale. Care-management is single-attempt
throughout; its bars are hatched because the baseline and procedure runs
used the now-retired sonnet-4 patient simulator while the advisory used
its successor sonnet-4-5 (a simulator-substitution caveat, not a metric
mismatch), and it is additionally confounded by simulated-member consent
variance. Source: Section 5.6.}
\end{figure}

Two metric distinctions are load-bearing and are stated plainly because
they bound every advisory number above. First, \emph{self-consistency is
not pass@1}: the 84\% and 44\% figures select the best of five
answer-blind draws, whereas the leaderboard publishes strict
single-attempt pass@1, which is lower (utilization-management 39.2\%;
prior-authorization 68\%, now confirmed by a held-out single-run board).
Second, the advisory injects case-specific policy grounding, which
raises a construct question the paper answers directly: is this
governance or teaching to the test? It is governance under one strict
condition, that the advisory reads only the case's own visible policy
and published standards and never any hidden benchmark surface, which
keeps it an instance of the framework's evidence-grounding and
epistemic-classification commitments rather than answer injection. Where
that line is crossed it stops being governance, so the answer-blind
constraint is the whole methodological point, not a footnote.

Care-management is where the advisory line meets its wall, and the
honest account of that domain is a sequence of corrections. Grounding
the care-management advisory in the live United States Medicare
care-management programs (chronic care management, advanced primary care
management, and the dementia GUIDE model) reached \textbf{40\%
single-draw pass@1 (10 of 25, mean fractional 0.896)}. This makes
care-management the one domain where the generic procedure (the erius
submission's 56\%) outscores the case-grounded content advisory, the
inverse of prior-authorization and utilization-management, subject to
the simulator caveat in the Table 6 note: content quality is graded by a
subjective judge, and a definition-of-done cannot specify its way past
that judge. A further round of sharpening reached \textbf{52\% (13 of
25)} but as redistribution, not clean lift: it crossed five tasks and
clipped two previously passing ones, and a subsequent answer-blind
trajectory diagnosis overturned the round's own headline. The one
apparent structural crossing was not the lever working; in that draw the
simulated member happened to consent, so the case ran the ordinary
pipeline. This exposed the domain's governing confound: because a
nominally refusing member sometimes consents and sometimes refuses
across draws, and the agent's entire downstream pipeline turns on that,
much of the single-draw movement on care-management refusal tasks is
simulated-member behavior variance rather than advisory efficacy, and
isolating the advisory's effect requires reading each draw's consent
state or running more than one draw. The two structural floors were, on
diagnosis, genuine recording gaps (the agent declined to fabricate a
plan, correctly, but then closed via an intake shortcut instead of
persisting the pipeline's terminal record), and the two regressions were
genuine over-scaffolding artifacts (a forced-false social-determinant
field contradicting its own narrative; the full assessment battery run
inside the outreach call to the turn limit). The nine remaining
near-misses are the content-quality wall proper: subjective
large-language-model-judge grades that three independent rounds (a
separated content-critic, the grounding round, and the sharpening round)
failed to move, consistent with the paper's Finding 4 judgment-ceiling
account.

Finally, the advisory work was partly automated. An advisory-space
recursive-self-improvement loop (grade the board, read the failing
checks, refine the answer-blind facts, A/B-gate the change, keep or
discard) re-derived the hand-found fixes from the failure signal alone
and then correctly declined to churn the already-sharp facts, mapping
precisely where answer-blind advisory plateaus. This matters for the
framework because it optimizes the \emph{advisory}, not the model
weights: the plateau it finds is the boundary of what
governance-artifact grounding can do without either a different verifier
or a capability change through training, and it delineates that boundary
rather than crossing it.

The advisory approach also replicates across models, not only on Opus.
On a weaker frontier model, Google's Gemini 3.6 Flash (run through its
native command-line harness on a subscription, answer-blind,
single-attempt), the same pipeline lifts prior-authorization from a 12\%
unassisted baseline (3 of 25) to 20\% with the generic advisory (5 of
25), and then, through two rounds of the answer-blind
recursive-self-improvement loop described above (a site-of-service
redirect grounded in the ambulatory-surgical-center place-of-service
policy, then a chart-documentation-fidelity diagnosis-code correction),
to 32\% (8 of 25). The absolute lift is smaller than Opus's and the
residual is the same structural wall (eleven disposition-judgment tasks
plus a few model-floor cases), but the direction reproduces: a
case-grounded, answer-blind definition-of-done raises a lower-capability
frontier model on the same recoverable dimensions it raises Opus. Two
caveats bound the result: the figures are single-attempt draws on a
high-variance host path, and the recursive-self-improvement rounds were
developed against the same 25 public tasks, so, as with the Opus
advisory, a fresh-task evaluation is the stronger generalization test.
This Gemini 3.6 Flash arm is distinct from the Gemini 3.1 Pro procedure
arm of Section 5.4.

The advisory also reproduces across a model GENERATION, not only across
vendors. Re-running the full three-domain board on Opus 5 (Anthropic's
next-generation frontier model) under the same single-agent,
single-attempt protocol (reasoning-effort settings tuned per domain:
default on prior-authorization, higher on utilization-management and
care-management) gives prior-authorization 18 of 25 (72\%),
utilization-management 9 of 25 (36\%), and care-management 14 of 25
(56\%), for 41 of 75 (54.7\%) overall. The pattern from Opus 4.8 holds
on the newer model: prior-authorization is led by the answer-blind
advisory with the same chart-documentation-fidelity corrections (a
place-of-service and a diagnosis-code fix, research-grounded and never
read from the key), and care-management, run procedure-only (the generic
procedure under the same agent, with no content advisory, on the
sonnet-4-5 patient simulator), lands at 56\%, matching Opus 4.8's erius
care-management exactly. No content-advisory arm was run on
care-management for Opus 5, so this reproduces the procedure's
care-management level on a second model generation; it is consistent
with, but does not by itself re-test, the procedure-beats-advisory
inversion documented above. Utilization-management on Opus 5 settles at
36\% under a higher reasoning-effort setting; a terminal sign-off commit
lever tested there was rejected under the study's zero-regression rule
(it recovered one sign-off task but regressed a passing one, net zero),
consistent with the Section 5.5 finding that the sign-off boundary is a
judgment ceiling a prompt lever redistributes rather than crosses. All
Opus 5 figures are strict single-attempt pass@1, directly comparable to
the leaderboard estimand, though not filed as a submission; as with the
other replication arms, the board runs the same 25 public tasks per
domain with a single trial each, so a fresh-task evaluation remains the
stronger generalization test.

To keep the comparison honest, the arms are set side by side on ONE
estimand, single-attempt pass@1 (the metric the leaderboard scores), and
the advisory's best-of-five self-consistency selection is reported
separately as a ceiling on the same draws, not as a single-attempt
result.

\textbf{Configuration ladder (each column is one rung; model and harness
held constant, one layer added at a time).} \emph{Baseline (no GAMPO):}
the unscaffolded tool-use loop, no procedure and no advisory.
\emph{Procedure:} the same loop with the generic GAMPO operating
procedure prepended, uniform governance with no case-specific content.
\emph{Advisory, single-attempt:} the procedure plus generic
decision-discipline modules plus an answer-blind, per-task
definition-of-done keyed only to the case's own visible policy and
published United States standards and never the hidden key, scored on
one draw (the leaderboard estimand). \emph{Advisory, best-of-five
ceiling:} the same advisory under answer-blind self-consistency
selection over five draws, an upper ceiling and not a single-attempt
number. A fifth, facts-ablated arm (procedure plus generic
decision-discipline modules with the per-task facts removed) is the
held-out generalization test that isolates how much of the advisory lift
is the case-specific definition-of-done versus the generic governance
that ports without tuning; it reaches 48\% single-attempt (versus 68\%
full and 24\% baseline), so the generic governance generalizes and the
facts add a smaller increment (Section 5.7).

\begin{longtable}[]{@{}
  >{\raggedright\arraybackslash}p{(\linewidth - 8\tabcolsep) * \real{0.2202}}
  >{\centering\arraybackslash}p{(\linewidth - 8\tabcolsep) * \real{0.1193}}
  >{\centering\arraybackslash}p{(\linewidth - 8\tabcolsep) * \real{0.1560}}
  >{\centering\arraybackslash}p{(\linewidth - 8\tabcolsep) * \real{0.2385}}
  >{\centering\arraybackslash}p{(\linewidth - 8\tabcolsep) * \real{0.2661}}@{}}
\caption{Governance progression on Opus 4.8 by domain: single-attempt
pass@1 (the leaderboard estimand) with the advisory's best-of-five
self-consistency selection as a separate ceiling; Wilson 95\% CIs;
source Sections R24--R36.}\tabularnewline
\toprule\noalign{}
\begin{minipage}[b]{\linewidth}\raggedright
Domain
\end{minipage} & \begin{minipage}[b]{\linewidth}\centering
Baseline
\end{minipage} & \begin{minipage}[b]{\linewidth}\centering
Procedure
\end{minipage} & \begin{minipage}[b]{\linewidth}\centering
Advisory, single-attempt
\end{minipage} & \begin{minipage}[b]{\linewidth}\centering
Advisory, best-of-5 ceiling
\end{minipage} \\
\midrule\noalign{}
\endfirsthead
\toprule\noalign{}
\begin{minipage}[b]{\linewidth}\raggedright
Domain
\end{minipage} & \begin{minipage}[b]{\linewidth}\centering
Baseline
\end{minipage} & \begin{minipage}[b]{\linewidth}\centering
Procedure
\end{minipage} & \begin{minipage}[b]{\linewidth}\centering
Advisory, single-attempt
\end{minipage} & \begin{minipage}[b]{\linewidth}\centering
Advisory, best-of-5 ceiling
\end{minipage} \\
\midrule\noalign{}
\endhead
\bottomrule\noalign{}
\endlastfoot
Prior-authorization & 24\% {[}12, 43{]} & 40\% {[}23,59{]} & 68\%
(17/25) {[}48,83{]}* & 84\% (21/25) {[}65,94{]} \\
Utilization-management & 8\% {[}2,25{]} & 16\% {[}6,35{]} & 39.2\%** &
44\% (11/25) {[}27,63{]} \\
Care-management & 40\% {[}23,59{]} & 56\% (14/25) {[}37,73{]} & 40\%
(10/25) {[}23, 59{]} & n/a (single-draw) \\
\end{longtable}

Table Note. Every column is single-attempt pass@1 except the last, which
reports the advisory's best-of-five answer-blind self-consistency
selection over the same five draws, an upper ceiling, not a
single-attempt number. *The prior-authorization single-attempt figure is
now CONFIRMED by a dedicated single-run board (the frozen advisory, one
draw per task, Opus 4.8 on the subscription, run held-out): \textbf{68\%
(17 of 25), Wilson 95\% {[}48, 83{]}}, which lands inside the earlier
high-60s-to-low-70s estimate. The 16-point gap to the 84\% best-of-five
ceiling is the self-consistency premium measured on the same draws; the
eight single-attempt misses are the self-consistency-dependent tasks.
**The utilization-management single-attempt figure IS confirmed by a
single-run board and equals the mean binary pass over five draws per
task across 25 tasks (equivalently 9.8 effective passes / 25 = 39.2\%).
Care-management: both the erius 56\% (14/25) and the grounded advisory
40\% (10/25) are single-attempt but ran on different patient simulators
(the erius on the now-retired sonnet-4, the advisory on its successor
sonnet-4-5), so the comparison carries a simulator-substitution caveat.
A sharpened advisory round reached 52\% (13/25) single-draw but was
overturned by an answer-blind trajectory diagnosis as redistribution
under simulated-member consent variance, so it is not treated as the
advisory's result. On the numbers as filed, care-management is the one
domain where the generic procedure outscores the case-grounded content
advisory, the inverse of prior-authorization and utilization-management.
All advisory numbers are exploratory and, for care-management,
additionally confounded by simulated-member consent variance. The
care-management baseline (40\%) in the baseline column is the official
public-board figure for the same model; the local-harness
care-management baseline was invalidated by the dead-simulator artifact
(Section 5.7), so the baseline column mixes in-harness
(prior-authorization, utilization-management) and official-board
(care-management) provenance.

\subsection{5.7 Robustness and
sensitivity}\label{robustness-and-sensitivity}

Three checks bound the claims. \emph{Single-trial variance:} each task
was run once per arm; a re-roll flipped 2 of 5 borderline tasks and two
frontier runs disagreed on absolute level (40\% vs.~32\%) while agreeing
on the paired lift (+16 points both), so absolute single-run levels
overstate stable differences and the paired \emph{lifts} are the durable
signal. \emph{Care-management confound (resolved) and the
non-generalizing six-task lift:} an empty simulator credential initially
floored all local care-management trials at about 0.273, an artifact;
with a funded credential a six-task paired frontier A/B showed baseline
about 0.94 and governance nominally doubling pass@1 (2/6 to 4/6, paired
Delta +0.018). That six-task lift does not generalize: full 25-task
care-management A/Bs on two newer models fail to replicate it
(GPT-5.6-sol binary passes fall from 6/24 to 2/24 under GAMPO,
fractional Delta +0.007; Kimi 2.7-code fractional Delta -0.188), so we
treat it as small-sample optimism and report care-management as
content-quality-walled where comprehensive governance is
neutral-to-negative (Section 5.5). The earlier dead-simulator floor
readings (about 0.273) remain superseded and should not be cited.
\emph{Two further care-management threats surfaced in the advisory work
(Section 5.6) and bound its numbers.} First, a
\emph{benchmark-fragility} issue: the benchmark's pinned
patient-simulator model has been retired from its vendor's application
programming interface, so a fresh care-management run today floors at
the dead-simulator artifact unless the simulator is repointed to a
current model; the advisory boards used the nearest living successor, a
substitution that is necessary but leaves care-management figures not
strictly comparable to the frozen baseline until the upstream pin is
updated. Second, a \emph{consent-variance confound}: because the
simulated member's decision to consent or refuse varies across draws on
the refusal tasks, and the agent's downstream pipeline (and therefore
its pass or fail) turns on that decision, single-draw care-management
movement on those tasks conflates advisory effect with member-behavior
variance, so those results require consent-controlled attribution or
multiple draws and are reported as exploratory. \emph{Calibration:} the
harness under-scores one specific frontier model versus its public
figures but reproduces another model's official number exactly, so
within-study governance-versus-baseline contrasts (which hold model and
harness constant) are unaffected even where absolute cross-tier levels
are not directly comparable.

\emph{Held-out ablation (governance versus benchmark tuning), the
confirmatory step.} Both reviewers named the same top threat to the
advisory result: because the per-task facts were refined against graded
failures of the same 25 prior-authorization tasks, the 84\% could be
adaptive benchmark optimization rather than governance. A
pre-registered, frozen, held-out run separates the two. With the
advisory modules and per-task facts frozen and the runtime pinned, and
scored once per task under the single-attempt estimand, the full
advisory (procedure plus generic decision-discipline modules plus
per-task facts) reaches \textbf{68\% (17/25, Wilson 95\% {[}48, 83{]})}
and a facts-ablated arm (procedure plus the generic decision-discipline
modules, per-task facts emptied) reaches \textbf{48\% (12/25, {[}30,
67{]})}. The facts-ablated arm's Wilson lower bound, 30\%, clears the
24\% no-GAMPO baseline, so the generic governance generalizes: with no
case-specific facts at all, the procedure-and-modules kernel doubles the
unassisted baseline (24\% to 48\%). The per-task facts then add a
further paired +5 tasks (7 tasks the facts recover against 2 they cost
through redistribution; continuity-corrected McNemar chi-square 1.78
over nine discordant pairs, underpowered), a directional increment not
statistically resolved at n = 25. The honest reading is that the
advisory's lift is mostly generalizable governance rather than benchmark
tuning, with a smaller case-specific facts increment a larger sample
would be needed to confirm. This closes the confirmatory step flagged in
Future Work item 2 for prior-authorization; the equivalent held-out on
fresh tasks outside the public 25 remains the stronger future test. A
cross-generation check points the same way: the advisory's
prior-authorization result reproduces on Opus 5 at 72\% single-attempt
(18/25), numerically above the Opus 4.8 held-out 68\% (a one-task
difference at n = 25, well inside the overlapping Wilson intervals);
utilization-management lands at 36\% (9/25), slightly below the Opus 4.8
single-attempt 39.2\%; and the care-management procedure result
reproduces at 56\% (14/25) exactly, on the successor sonnet-4-5
simulator. The governance effect is therefore not idiosyncratic to a
single model version, though the check runs the same 25 public tasks
with a single trial per task.

\section{6. Discussion}\label{discussion}

\textbf{Interpretation.} The unifying explanatory hypothesis is
\emph{spare capacity} (supported, not established as causal). A
governance procedure is an instruction the model must hold in working
memory and act on while also doing the task; when capacity is scarce,
the procedure competes with the work and can crowd it out, which is why
a 35B model is hurt by the full framework but helped by a single
verification sentence, and why a frontier model with headroom can spend
the procedure productively. This reframes governance scaffolding from a
uniform good to a capacity-dependent intervention, and it predicts the
otherwise puzzling pattern that the cheapest, most targeted intervention
(verify your writes) was the strongest arm on the constrained model.
Where governance did not help the frontier, the binding constraint was
the model's judgment disposition, not the scaffold, which is why prompt
levers mostly failed but a \emph{bias-removing} rewrite succeeded where
a \emph{permission-granting} one did not.

\textbf{Implications for executives.} Govern in proportion to spare
capacity. For the capacity-constrained models that dominate
cost-sensitive production, prefer minimal, action-grounded verification
over comprehensive procedure, and expect comprehensive governance
scaffolds to pay off only on frontier models, and even there only on
domains where the model already has traction. Do not buy a uniform
``governance lift''; budget for domain- and model-specific evaluation.
Retain humans at non-delegable boundaries regardless of model
capability, because the failure modes that governance cannot fix
(deference on a clinical override, a status string that penalizes a
clinically correct action) are exactly the ones with the highest cost of
error.

\textbf{Implications across people, process, technology, and economics.}
Read through the people-process-technology-economics lens of the
companion prior-authorization readiness study (Johnson, 2027b), the
findings translate into four organizational implications (Table 5); the
people and economic rows are reasoned from the results rather than
directly measured.

\begin{longtable}[]{@{}
  >{\raggedright\arraybackslash}p{(\linewidth - 4\tabcolsep) * \real{0.1277}}
  >{\raggedright\arraybackslash}p{(\linewidth - 4\tabcolsep) * \real{0.4255}}
  >{\raggedright\arraybackslash}p{(\linewidth - 4\tabcolsep) * \real{0.4468}}@{}}
\caption{People, process, technology, and economic implications of the
capability-gated findings (source Sections 5.2--5.6; people and economic
rows are derived implications, not measured effects).}\tabularnewline
\toprule\noalign{}
\begin{minipage}[b]{\linewidth}\raggedright
Dimension
\end{minipage} & \begin{minipage}[b]{\linewidth}\raggedright
From the study
\end{minipage} & \begin{minipage}[b]{\linewidth}\raggedright
Implication
\end{minipage} \\
\midrule\noalign{}
\endfirsthead
\toprule\noalign{}
\begin{minipage}[b]{\linewidth}\raggedright
Dimension
\end{minipage} & \begin{minipage}[b]{\linewidth}\raggedright
From the study
\end{minipage} & \begin{minipage}[b]{\linewidth}\raggedright
Implication
\end{minipage} \\
\midrule\noalign{}
\endhead
\bottomrule\noalign{}
\endlastfoot
People & Judgment-ceiling and deference failures governance cannot fix
(clinical override, sign-off) & Keep humans on the non-delegable
determinations; scaffolding does not substitute for judgment \\
Process & Govern in proportion to spare capacity; a one-sentence
verification beats the full procedure & Make governance
capacity-proportional; default to a lightweight verification control,
not a comprehensive scaffold \\
Technology & Benefit is gated by model tier, not the scaffold; only the
prompt-layer kernel (8 of 26 components) was instrumented & Model
selection dominates scaffolding for governed workflows; choose models
with spare capacity; the full runtime is now implemented (Figure 12) but
its ungated components await end-to-end evaluation \\
Economics & A one-sentence prompt outperforms a 5,200-character
procedure on weak models; the filed submission tops out at 37.3\% &
Govern at the token cost the model can absorb; the benchmark quantifies
today's automation ceiling for these workflows \\
\end{longtable}

\textbf{From specification to runnable implementation.} The framework is
not only a specification. A runnable sibling repository
(\texttt{gampo-runtime}) now instantiates the architecture, and Figure
12 maps each C4 and Appendix-C element to its implementation status. The
reasoning and governance core is fully realized: the ten-capability
autotelic cycle (as nine Model Context Protocol capability servers, with
goal prioritization folded into the filtering server), the
constraint-and-promotion gate, the answer-blind governed retrieval, the
self-consistency selector, and the append-only audit ledger. The
integration surfaces (FHIR Prior-Authorization-Support, EHR-embedded
middleware, the payer-provider bridge) are answer-blind simulations, and
live CI/CD and production observability remain stubbed. This
distinguishes what the runtime \emph{implements} from what the benchmark
probe \emph{exercises}: the empirical findings above still bound only
the prompt-layer kernel (Figure 5), so what remains future work (item 1)
is the end-to-end evaluation of the runtime's ungated components, not
their construction.

\begin{figure}
\centering
\includegraphics[width=0.95\linewidth,height=0.78\textheight,keepaspectratio,alt={Figure 12. GAMPO framework architecture mapped to the gampo-runtime implementation: each C4 and Appendix-C element against its status in the runnable sibling repository and the CHI-Bench harness family. The reasoning and governance core is fully instantiated; integration surfaces are answer-blind simulations (a FHIR Prior-Authorization-Support simulator, a frozen criteria corpus); live EHR and payer connectivity, CI/CD, and production observability are deferred. Source: gampo-runtime repository audit.}]{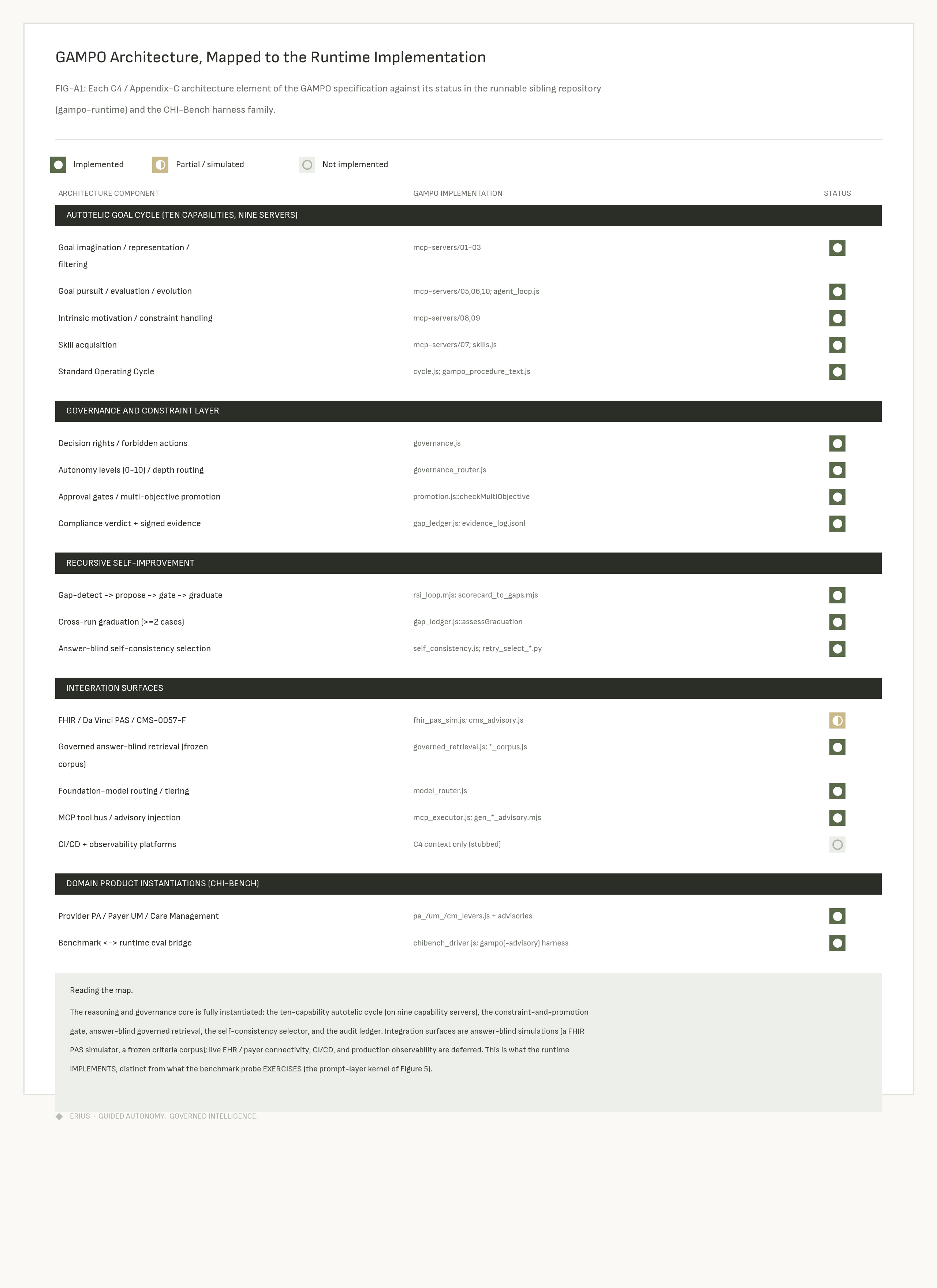}
\caption{GAMPO framework architecture mapped to the
gampo-runtime implementation: each C4 and Appendix-C element against its
status in the runnable sibling repository and the CHI-Bench harness
family. The reasoning and governance core is fully instantiated;
integration surfaces are answer-blind simulations (a FHIR
Prior-Authorization-Support simulator, a frozen criteria corpus); live
EHR and payer connectivity, CI/CD, and production observability are
deferred. Source: gampo-runtime repository audit.}
\end{figure}

Read against the companion prior-authorization readiness study's
architecture-and-integration pattern catalog (Johnson, 2027b), whose
thirteen coded patterns and seven-layer reference architecture define
what scale-ready prior-authorization deployments most consistently
report, GAMPO covers seven of the thirteen patterns fully and five as
answer-blind simulations, with one (a legacy non-FHIR bridge) out of
scope (Figure 13). This coverage profile mirrors the study's own
principal Technology finding: architecture is the most evidentially
mature readiness dimension, while a minimum viable architecture for live
regulated deployment, the integration and continuous-monitoring surfaces
GAMPO leaves simulated, remains absent.

\begin{figure}
\centering
\includegraphics[width=0.95\linewidth,height=0.78\textheight,keepaspectratio,alt={Figure 13. GAMPO coverage of the companion prior-authorization readiness study's scale-readiness pattern catalog (Johnson, 2027b): the thirteen coded agentic-architecture, integration-architecture, and technical-control patterns (RQ1a) and the seven-layer reference architecture, each mapped to its GAMPO instantiation and status. Seven of thirteen patterns are fully instantiated (the agentic and control core); five are answer-blind simulations (the healthcare-interoperability and live-human or drift surfaces); one is out of scope. Source: author's mapping of the companion catalog to the gampo-runtime implementation.}]{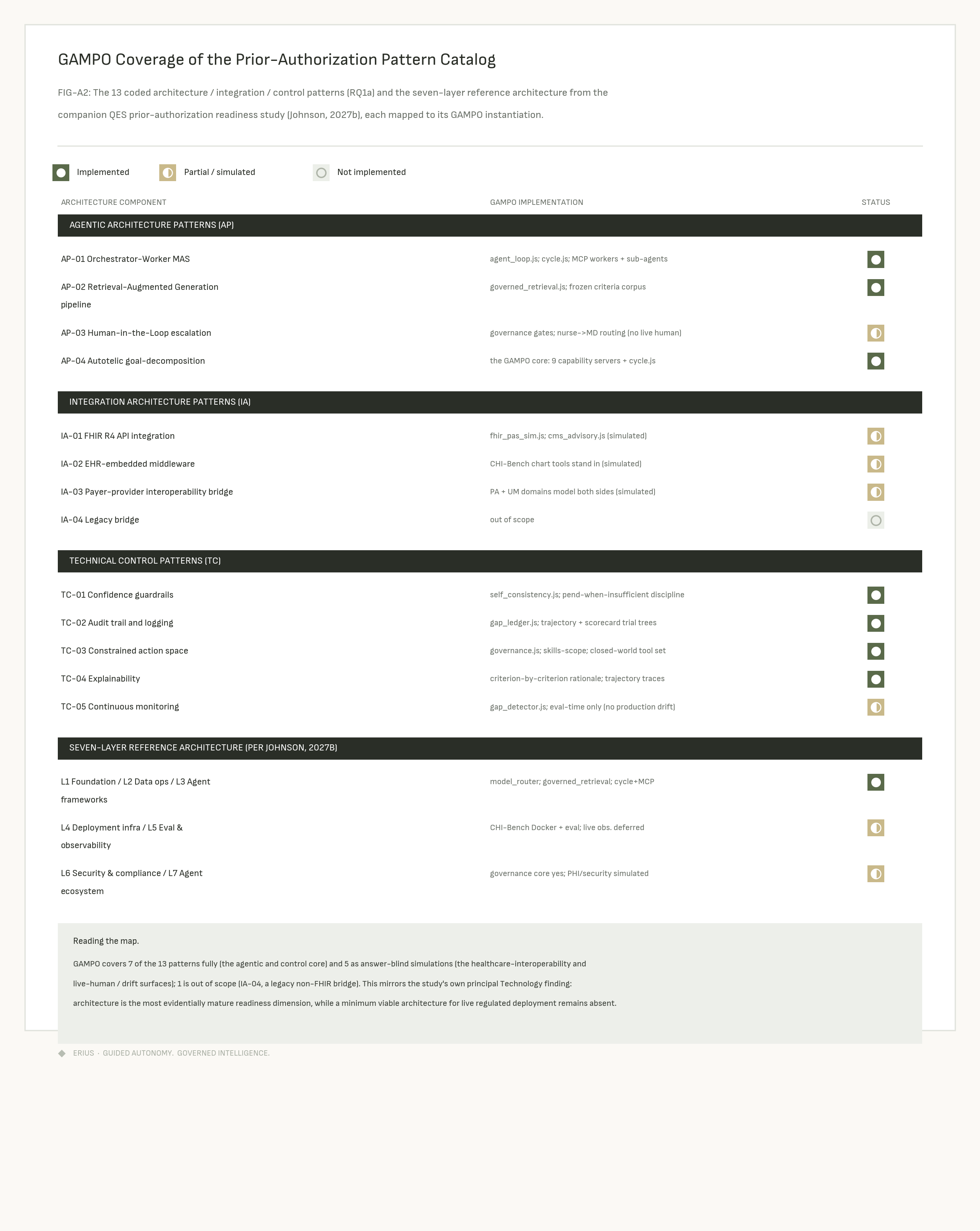}
\caption{GAMPO coverage of the companion prior-authorization
readiness study's scale-readiness pattern catalog (Johnson, 2027b): the
thirteen coded agentic-architecture, integration-architecture, and
technical-control patterns (RQ1a) and the seven-layer reference
architecture, each mapped to its GAMPO instantiation and status. Seven
of thirteen patterns are fully instantiated (the agentic and control
core); five are answer-blind simulations (the
healthcare-interoperability and live-human or drift surfaces); one is
out of scope. Source: author's mapping of the companion catalog to the
gampo-runtime implementation.}
\end{figure}

The catalog is not only a coverage checklist: an accuracy-lever subset
of these patterns is what drove the benchmark result upward, and
separating the score-moving patterns from the readiness and governance
patterns is itself informative. Figure 14 attributes each measured
stage-lift in the Opus 4.8 prior-authorization progression (24\% to 84\%
pass@1) to the Figure-13 pattern(s) it operationalizes. The autotelic
procedure and constrained action space (AP-04, TC-03) add the first
sixteen points; the generic decision-discipline modules with governed
retrieval, criterion-by-criterion explainability, and the
pend-when-insufficient guardrail (AP-02, TC-04, TC-01) add eight on a
held-out board; per-task chart-documentation-fidelity grounding (AP-02)
adds the largest board increment (+20 board, +5 net paired after
redistribution); and answer-blind self-consistency selection (TC-01)
adds the final sixteen under a best-of-five estimand rather than the
single-attempt metric the leaderboard scores. The remaining patterns
(orchestrator-worker structure, FHIR and EHR interoperability, the
payer-provider and legacy bridges, audit logging, continuous monitoring)
enable deployment but do not move pass@1. These attributions map
measured stage lifts to the patterns they operationalize; they are not
an isolated thirteen-way ablation, and the human-in-the-loop deference
repair (AP-03, Section 5.5) acted on specific intervention-resistant
tasks rather than a whole stage.

\begin{figure}
\centering
\includegraphics[width=0.95\linewidth,height=0.78\textheight,keepaspectratio,alt={Figure 14. How the pattern catalog raised the prior-authorization score: the Opus 4.8 prior-authorization pass@1 built up stage by stage (24\% to 84\%), each measured lift attributed to the Figure-13 pattern(s) it operationalizes (baseline to procedure, AP-04 and TC-03; decision-discipline modules, AP-02, TC-04, TC-01; per-task chart-fidelity facts, AP-02; self-consistency selection, TC-01). The +20 facts increment is a board delta (+5 net paired after redistribution) and the final +16 is best-of-five, not single-attempt. The readiness and governance patterns (orchestrator-worker, FHIR/EHR interoperability, audit logging, monitoring) enable deployment but are not pass@1 levers. Attributions map measured stage lifts to patterns, not an isolated ablation. Source: Sections 5.2, 5.6, and 5.7.}]{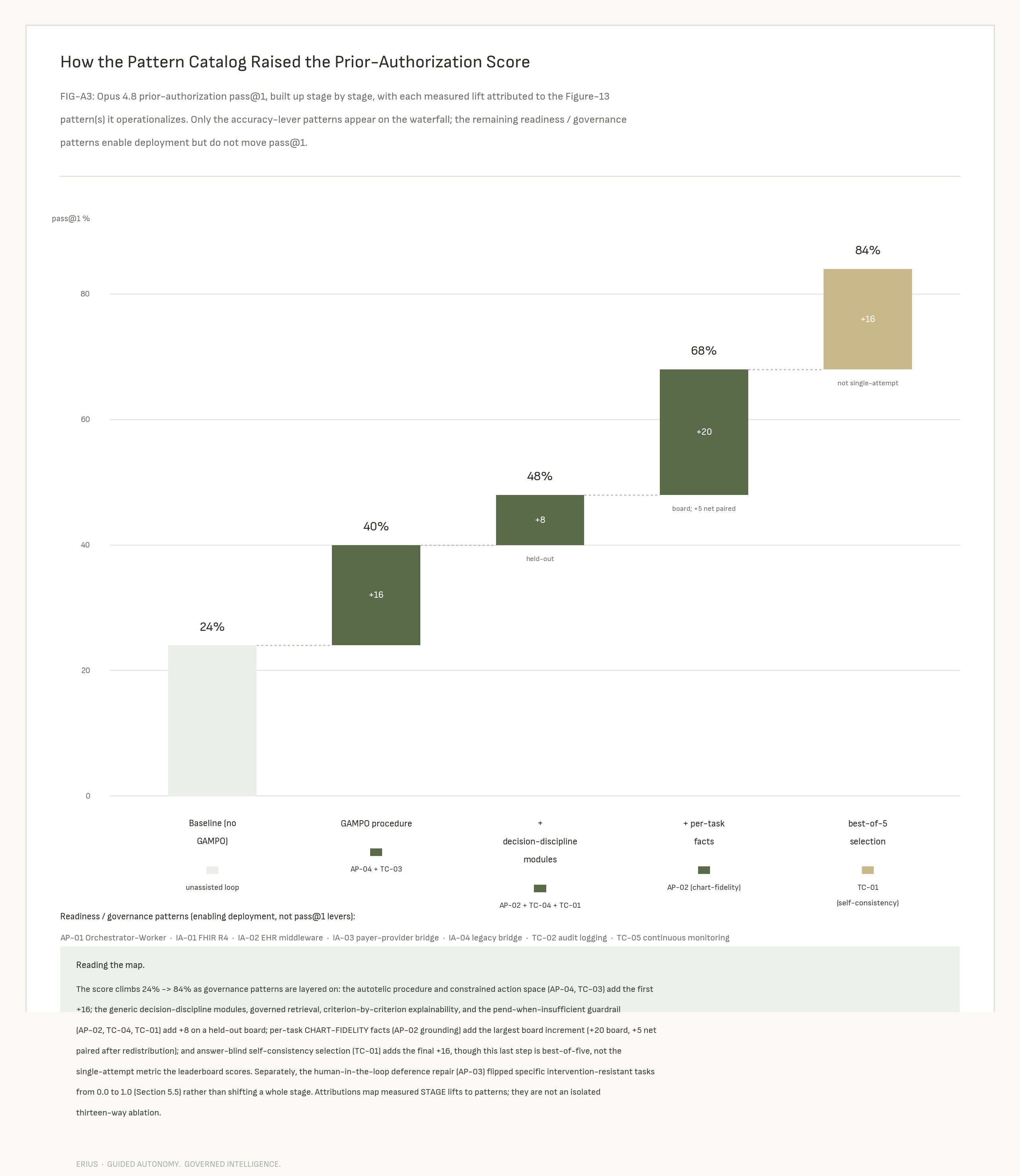}
\caption{How the pattern catalog raised the
prior-authorization score: the Opus 4.8 prior-authorization pass@1 built
up stage by stage (24\% to 84\%), each measured lift attributed to the
Figure-13 pattern(s) it operationalizes (baseline to procedure, AP-04
and TC-03; decision-discipline modules, AP-02, TC-04, TC-01; per-task
chart-fidelity facts, AP-02; self-consistency selection, TC-01). The +20
facts increment is a board delta (+5 net paired after redistribution)
and the final +16 is best-of-five, not single-attempt. The readiness and
governance patterns (orchestrator-worker, FHIR/EHR interoperability,
audit logging, monitoring) enable deployment but are not pass@1 levers.
Attributions map measured stage lifts to patterns, not an isolated
ablation. Source: Sections 5.2, 5.6, and 5.7.}
\end{figure}

\textbf{Implications for academics.} Methodologically, the study shows
that QES plus a single, well-instrumented prompt-layer probe can both
produce an auditable framework and falsify the easy version of its
central claim, a useful template for applied-technology doctoral work in
fast-churning domains. Substantively, it contributes a capability-gated
account of agent governance and a concrete, replicated explanatory
hypothesis (spare capacity) that future benchmarks can test directly;
the deference result also isolates a measurable, model-level disposition
(recommendation override) that is partly prompt-addressable, inviting
targeted study.

\textbf{Boundaries.} The empirical findings concern a prompt-layer
kernel of the framework on one healthcare benchmark; they should not be
read as evaluating the full GAMPO specification, nor as generalizing
beyond long-horizon, policy-rich tool-use tasks. The framework's broader
propositions (topology leanness, the five-layer evaluation, the evidence
ledger) remain synthesis-derived and untested here.

\section{7. Limitations and Future
Research}\label{limitations-and-future-research}

\textbf{Limitations.} \emph{Design:} the framework synthesis collects no
primary data and its conclusions are confined to the 321-source
population, with the usual single-coder, publication-bias, and
English-language constraints. \emph{Construct:} the benchmark probe
instruments 8 of 26 components (4 partially), so it tests the
prompt-layer kernel, not the full framework; the ledger, coercive
policy-as-code, human-approval gates, and five-layer evaluation,
although now implemented in the gampo-runtime repository (Figure 12),
are not exercised by the probe. \emph{Statistical:} per-cell n = 5--25
and a single trial per task make every benchmark result exploratory; the
headline verify-base advantage is descriptively large but not
significant under a conservative paired test, and absolute single-run
levels carry draw-to-draw variance. \emph{Procedural:} the mandate
rebalance was folded into the default procedure mid-study, so some
earlier rows used prior wording (valid as recorded, but not reproducible
from the current module without reverting), and it has not been
regression-checked for over-denial on approve-correct tasks.
\emph{External:} open-model and frontier arms used different harnesses,
so cross-tier comparisons are directional.

\textbf{Future work.} (1) Evaluate \texttt{gampo-runtime} end-to-end:
the full instantiation now exists (Figure 12), with the reasoning and
governance core implemented and integration surfaces simulated, so what
remains is to exercise the ledger, coercive policy-as-code,
human-approval gates, and five-layer evaluation under load, the
components the prompt-layer probe could not reach (addresses the
construct limitation), and to replace the simulated FHIR and payer
surfaces with live connections. (2) Power the key contrasts:
multi-trial, larger-n A/Bs with pre-registered paired tests to convert
the verify-base and frontier-lift signals from exploratory to
confirmatory, and, for the answer-blind advisory specifically, extending
the completed frozen held-out evaluation (Section 5.7: 68\% full, 48\%
facts-ablated on prior-authorization) to fresh tasks outside the public
25 and to the other two domains, to further separate governance efficacy
from adaptive benchmark optimization. (3) A full approve-side regression
of the mandate rebalance to bound its over-denial risk before
deployment. (4) A capacity-graded study that manipulates spare capacity
directly (effort budgets, model size) to test the spare-capacity
mechanism as a causal claim rather than an inference. (5) Replication on
a non-healthcare long-horizon benchmark to test domain generality of the
capability-gating result. (6) Run the governance probe on a
healthcare-specialized, tool-capable frontier model rather than a
general one: Cura 1T (Chen et al., 2026b), from the benchmark's own
authors, is trained through a human-gated self-evolution loop, a
training-time analogue of GAMPO's autotelic goal loop and of the
advisory-space recursive self-improvement in Section 5.6, and is a
direct test of whether the capability-gating and governance-form
findings hold when the model is both domain-specialized and agentic (in
contrast to the medically fine-tuned model that this study had to
exclude for lacking tool-calling). (7) Extend the open-model
capability-gate cohort with additional open-weight lineages, notably
Nous Research's Hermes family, to test whether the spare-capacity
boundary and the one-sentence verify-base effect reproduce across model
families this probe did not include, and to widen the range over which
the spare-capacity mechanism is sampled. (8) Test harness-invariance:
re-run the governance probe under an alternative agentic framework,
OpenClaw, rather than this study's shared OpenAI-Agents tool-use loop
and the frontier models' native command-line harnesses, to confirm that
the capability-gating and governance-form findings are properties of
model and task rather than artifacts of the specific agent harness (a
robustness check the single-harness design cannot itself provide).

\section{8. Conclusion}\label{conclusion}

This paper contributes a named, auditable governance framework for
autotelic multi-agent product organizations and capability-gated
evidence on whether instantiating such governance helps. Synthesizing
321 appraised sources under a pragmatist anchor, GAMPO specifies a
runnable, repository-based operating model, governance artifacts, an
epistemic claim taxonomy, nine non-delegable boundaries, a
ten-capability autotelic engine, and a five-layer evaluation. That
specification is now implemented in a runnable runtime whose reasoning
and governance core is complete and whose integration surfaces are
answer-blind simulations; the evidence here bounds only its prompt-layer
kernel. Probing its prompt-layer kernel on a long-horizon healthcare
benchmark then overturns the intuitive expectation that more governance
is uniformly better: governance benefit is gated by a model's spare
capacity and is domain- and model-specific. On capacity-constrained
models the full framework is at best neutral and a single ``verify your
writes'' sentence consistently wins in our runs (pooled fractional 0.431
to 0.590, pass@1 doubled); at the frontier the same scaffold lifts
paired success where the model has traction (prior-authorization 24\% to
40\%) yet not where its judgment disposition blocks it. A second
frontier result sharpens the rule: the \emph{form} of governance matters
as much as its presence, because replacing the uniform procedure with an
answer-blind, per-task definition-of-done, a case-grounded governance
artifact rather than a generic scaffold, roughly doubles the
prior-authorization result to 84\% under best-of-five self-consistency
(68\% single-attempt on a held-out board), while care-management meets a
subjective-judge content-quality wall that no governance-artifact round
crosses. The actionable takeaway is one rule with a number behind it:
\textbf{govern in proportion to spare capacity}, and for the constrained
models that run most production work, a one-sentence verification habit,
not a comprehensive procedure, is the highest-yield governance you can
add, while at the frontier the highest-yield governance is a
case-grounded specification, not a longer procedure.

\section{Disclosures}\label{disclosures}

\textbf{Source.} This paper is derived from the author's doctoral
dissertation proposal (Johnson, 2027a). The conceptual framework and its
evidence base come from that document-based synthesis; the empirical
benchmark study reported in Sections 5.2--5.6 is post-proposal work
conducted by the author and is not part of the dissertation's
document-based design.

\textbf{Ethics.} No human subjects were involved at any stage; the study
is non-human-subjects research under 45 CFR 46.

\textbf{Data and code availability.} The GAMPO agent submission (team
\emph{erius}) and its evaluation artifacts are available on the
CHI-Bench leaderboard at
\url{https://github.com/actava-ai/leaderboard/tree/main/benchmarks/chi-bench/submissions/2026-06-02-erius/}.

\textbf{Competing interests.} The author declares none.

\textbf{Benchmark.} The benchmark used for evaluation, CHI-Bench
(chi-Bench), is third-party work (Chen et al., 2026a).

\section{References}\label{references}

Ambler, S. W., \& Lines, M. (2022). \emph{Choose your WoW! A Disciplined
Agile approach to optimizing your way of working} (2nd ed.). Project
Management Institute.

Aromataris, E., \& Munn, Z. (Eds.). (2020). \emph{JBI manual for
evidence synthesis}. Joanna Briggs Institute.
\url{https://doi.org/10.46658/JBIMES-20-01}

Becker, J., Rush, N., Barnes, E., \& Rein, D. (2025). Measuring the
impact of early-2025 AI on experienced open-source developer
productivity. \emph{arXiv}. \url{https://arxiv.org/abs/2507.09089}

Bommasani, R., Hudson, D. A., Adeli, E., Altman, R., Arora, S., von Arx,
S., Bernstein, M. S., Bohg, J., Bosselut, A., Brunskill, E.,
Brynjolfsson, E., Buch, S., Card, D., Castellon, R., Chatterji, N.,
Chen, A., Creel, K., Davis, J. Q., Demszky, D., \ldots{} Liang, P.
(2021). On the opportunities and risks of foundation models.
\emph{arXiv}. \url{https://arxiv.org/abs/2108.07258}

Bornet, P., Wirtz, J., Davenport, T. H., De Cremer, D., Evergreen, B.,
Fersht, P., Gohel, R., Khiyara, S., Mullakara, N., \& Sund, P. (2025).
\emph{Agentic artificial intelligence: Harnessing AI agents to reinvent
business, work, and life}. World Scientific.

Brooks, F. P. (1975). \emph{The mythical man-month: Essays on software
engineering}. Addison-Wesley.

Chen, H., Metelski, D., Qi, L., Xia, T., Lee, J., Brown, S., Riley, K.,
Wang, F., Liu, T. Y. A., Capps, H., Tang, Z., Song, X., Kong, L., Feng,
F., Zeng, T., Liu, Z., Ma, Z., Jiang, H., Geng, F., \ldots{} Yao, W.
(2026a). CHI-Bench: Can AI agents automate end-to-end, long-horizon,
policy-rich healthcare workflows? \emph{arXiv}.
\url{https://arxiv.org/abs/2605.16679}

Chen, H., Qi, L., Brown, S., Metelski, D., Xia, T., Lee, J., Wang, Q.,
Riley, K., Wang, F., \& Yao, W. (2026b). Cura 1T: Specialized model for
agentic healthcare. \emph{arXiv}. \url{https://arxiv.org/abs/2607.15314}

Colas, C., Karch, T., Sigaud, O., \& Oudeyer, P.-Y. (2022). Autotelic
agents with intrinsically motivated goal-conditioned reinforcement
learning: A short survey. \emph{Journal of Artificial Intelligence
Research, 74}, 1159--1199. \url{https://doi.org/10.1613/jair.1.13554}

Critical Appraisal Skills Programme. (2018). \emph{CASP qualitative
studies checklist}. CASP. \url{https://casp-uk.net/casp-tools-checklists/}

Csikszentmihalyi, M. (1990). \emph{Flow: The psychology of optimal
experience}. Harper \& Row.

Dewey, J. (1938). \emph{Logic: The theory of inquiry}. Henry Holt.

European Commission High-Level Expert Group on Artificial Intelligence.
(2019). \emph{Ethics guidelines for trustworthy AI}. European
Commission.
\url{https://digital-strategy.ec.europa.eu/en/library/ethics-guidelines-trustworthy-ai}

Gaurav, S., Heikkonen, J., \& Chaudhary, J. (2025).
Governance-as-a-service: A multi-agent framework for AI system
compliance and policy enforcement. \emph{arXiv}.
\url{https://arxiv.org/abs/2508.18765}

Hevner, A. R., March, S. T., Park, J., \& Ram, S. (2004). Design science
in information systems research. \emph{MIS Quarterly, 28}(1), 75--105.
\url{https://doi.org/10.2307/25148625}

International Organization for Standardization/International
Electrotechnical Commission. (2023). \emph{Information technology --
Artificial intelligence -- Management system} (ISO/IEC 42001:2023). ISO.

Johnson, M. R. (2027a). \emph{Governed autotelic multi-agent product
organizations (GAMPO): A qualitative evidence synthesis of design,
governance, and evaluation} {[}Doctoral dissertation proposal, Purdue
University{]}. Purdue University Graduate School.

Johnson, M. R. (2027b). \emph{From promise to readiness: A PPTE evidence
synthesis of agentic AI readiness in prior authorization for large U.S.
integrated delivery networks and managed care organizations} {[}Doctoral
dissertation, Purdue University{]}. Purdue University Graduate School.

Morgan, D. L. (2014). Pragmatism as a paradigm for social research.
\emph{Qualitative Inquiry, 20}(8), 1045--1053.
\url{https://doi.org/10.1177/1077800413513733}

National Institute of Standards and Technology. (2023). \emph{Artificial
intelligence risk management framework (AI RMF 1.0)} (NIST AI 100-1).
U.S. Department of Commerce. \url{https://doi.org/10.6028/NIST.AI.100-1}

Nonaka, I., \& Takeuchi, H. (1995). \emph{The knowledge-creating
company: How Japanese companies create the dynamics of innovation}.
Oxford University Press.

Page, M. J., McKenzie, J. E., Bossuyt, P. M., Boutron, I., Hoffmann, T.
C., Mulrow, C. D., Shamseer, L., Tetzlaff, J. M., Akl, E. A., Brennan,
S. E., Chou, R., Glanville, J., Grimshaw, J. M., Hrobjartsson, A., Lalu,
M. M., Li, T., Loder, E. W., Mayo-Wilson, E., McDonald, S., \ldots{}
Moher, D. (2021). The PRISMA 2020 statement: An updated guideline for
reporting systematic reviews. \emph{BMJ, 372}, n71.
\url{https://doi.org/10.1136/bmj.n71}

Parker, G. G., Van Alstyne, M. W., \& Choudary, S. P. (2016).
\emph{Platform revolution: How networked markets are transforming the
economy and how to make them work for you}. W. W. Norton.

Peffers, K., Tuunanen, T., Rothenberger, M. A., \& Chatterjee, S.
(2007). A design science research methodology for information systems
research. \emph{Journal of Management Information Systems, 24}(3),
45--77. \url{https://doi.org/10.2753/MIS0742-1222240302}

Polanyi, M. (1966). \emph{The tacit dimension}. Doubleday.

Schank, M. (2023). \emph{Digital transformation success: Achieving
alignment and delivering results with the Process Inventory Framework}.
Apress.

Shea, B. J., Reeves, B. C., Wells, G., Thuku, M., Hamel, C., Moran, J.,
Moher, D., Tugwell, P., Welch, V., Kristjansson, E., \& Henry, D. A.
(2017). AMSTAR 2: A critical appraisal tool for systematic reviews.
\emph{BMJ, 358}, j4008. \url{https://doi.org/10.1136/bmj.j4008}

Thomas, J., \& Harden, A. (2008). Methods for the thematic synthesis of
qualitative research in systematic reviews. \emph{BMC Medical Research
Methodology, 8}, 45. \url{https://doi.org/10.1186/1471-2288-8-45}

Tiwana, A. (2014). \emph{Platform ecosystems: Aligning architecture,
governance, and strategy}. Morgan Kaufmann.

Tyndall, J. (2010). \emph{AACODS checklist}. Flinders University.
\url{https://dspace.flinders.edu.au/xmlui/handle/2328/3326}

Volkhover, A. (2026). \emph{Agentic spec-driven development: A practical
method for using AI to build complete specifications for software,
products, and knowledge work}. Independently published.

Whittemore, R., \& Knafl, K. (2005). The integrative review: Updated
methodology. \emph{Journal of Advanced Nursing, 52}(5), 546--553.
\url{https://doi.org/10.1111/j.1365-2648.2005.03621.x}

Wooldridge, M. (2009). \emph{An introduction to multiagent systems} (2nd
ed.). John Wiley \& Sons.

\end{document}